\documentclass[sn-vancouver,Numbered]{sn-jnl}

\usepackage{graphicx}%
\usepackage{multirow}%
\usepackage{amsmath,amssymb,amsfonts}%
\usepackage{amsthm}%
\usepackage{mathrsfs}%
\usepackage[title]{appendix}%
\usepackage{xcolor}%
\usepackage{textcomp}%
\usepackage{manyfoot}%
\usepackage{booktabs}%
\usepackage[ruled,noline]{algorithm2e}%
\usepackage{algpseudocode}%
\usepackage{listings}%
\usepackage{bbm}%
\usepackage{subcaption}%

\usepackage{fancyhdr}
\let\mc\multicolumn

\begin{document}

\title[Article Title]{BioBLP: A Modular Framework for Learning on Multimodal Biomedical Knowledge Graphs}

\author*[1,2,4]{\fnm{Daniel} \sur{Daza}}\email{d.dazacruz@vu.nl}
\author*[1,4]{\fnm{Dimitrios} \sur{Alivanistos}}\email{d.alivanistos@vu.nl}
\author[3]{\fnm{Payal} \sur{Mitra}}
\author[3]{\fnm{Thom} \sur{Pijnenburg}}
\author[1,4]{\fnm{Michael} \sur{Cochez}}
\author[2,4]{\fnm{Paul} \sur{Groth}}

\affil[1]{Vrije Universiteit Amsterdam}
\affil[2]{University of Amsterdam}
\affil[3]{\orgname{Elsevier B.V.}, \orgaddress{\city{Amsterdam}, \country{The Netherlands}}}
\affil[4]{\orgname{Discovery Lab, Elsevier}, \orgaddress{\city{Amsterdam}, \country{The Netherlands}}}

\abstract{\textbf{Background:} Knowledge graphs (KGs) are an important tool for representing complex relationships between entities in the biomedical domain.
Several methods have been proposed for learning embeddings that can be used to predict new links in such graphs.
Some methods ignore valuable attribute data associated with entities in biomedical KGs, such as protein sequences, or molecular graphs.
Other works incorporate such data, but assume that entities can be represented with the same data modality.
This is not always the case for biomedical KGs, where entities exhibit heterogeneous modalities that are central to their representation in the subject domain.

\textbf{Objective:} We propose a modular framework for learning embeddings in KGs with entity attributes, that allows encoding attribute data of different modalities while also supporting entities with missing attributes.
We additionally propose an efficient pretraining strategy for reducing the required training runtime.
We train models using a biomedical KG containing approximately 2 million triples, and evaluate the performance of the resulting entity embeddings on the tasks of link prediction, and drug-protein interaction prediction, comparing against methods that do not take attribute data into account.

\textbf{Results:} In the standard link prediction evaluation, the proposed method results in competitive, yet lower performance than baselines that do not use attribute data.
When evaluated in the task of drug-protein interaction prediction, the method compares favorably with the baselines.
We find settings involving low degree entities, which make up for a substantial amount of the set of entities in the KG, where our method outperforms the baselines.
We also observe that optimizing attribute encoders is a challenging task that increases optimization costs.
Our proposed pretraining strategy yields significantly higher performance while reducing the required training runtime.

\textbf{Conclusion:} Our proposed method addresses the challenge of embedding multi-modal KGs with a modular architecture that can encode a wide range of attributes, while supporting missing data for entities.
The method provides embeddings that are effective on various tasks relevant to the biomedical domain.
The results over low degree entities indicate the potential for improved performance in scientific discovery tasks, where understudied areas of the KG would benefit from link prediction methods.

Our implementation is available at \url{https://github.com/elsevier-AI-Lab/BioBLP}.
}


\maketitle

\section{Background}
\label{sec:background}

The domain of Life Sciences (LS) relies on large scale and diverse data; ranging from genomics and proteomics data, to electronic health care records, and adverse events reports~\cite{Ritchie2018,stephens2015big,Zhu2020}.

A critical challenge to effectively using this data and inferring complex relationships across different sources is that it is often not interoperable~\cite{Wilkinson2016} and it cannot be easily integrated, hindering the ability to query and analyze the available knowledge~\cite{Waagmeester2020}.
To address this problem, semantic technologies have been employed to create large scale LS Knowledge Graphs~\cite{belleau2008bio2rdf,williams2012open,Waagmeester2020,covid19kg2020,Himmelstein2017}.
Knowledge Graphs (KG) are graph structured knowledge bases of entities and their relations~\cite{kg-book}, enabling, for example, the study of the relationship between chemical compounds and secondary pharmacology~\cite{CHICHESTER2015399} as well as drug repurposing~\cite{Himmelstein2017}.
Examples of KGs built for the LS include Bio2RDF~\cite{belleau2008bio2rdf}, Open PHACTS~\cite{williams2012open}, and Hetionet~\cite{Himmelstein2017}, which represent extensive repositories of various relationships between entities in the biomedical domain, as well as descriptive attributes about entities such as molecular structures, amino acid sequences, and disease descriptions.
This information can be complemented with additional data from resources such as DrugBank~\cite{knox2010drugbank}, UniProt~\cite{uniprot2015uniprot}, and MeSH~\cite{lipscomb2000medical}.

\begin{figure}[t]
    \centering
    \includegraphics[width=0.7\textwidth]{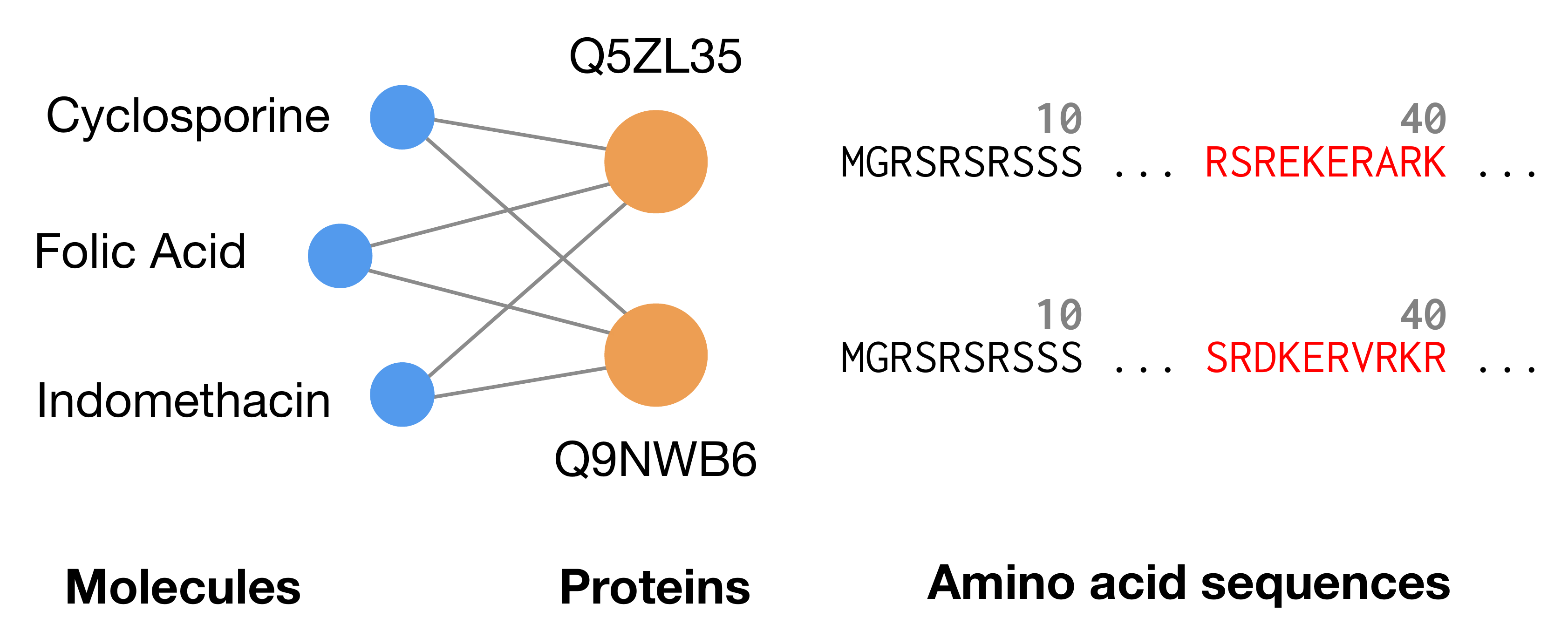}
    \caption{Two proteins (labeled with UniProt identifiers) from the BioKG dataset, which interact with the same molecules. 
When viewed from the perspective of the edges in the knowledge graph, the proteins are indistinguishable from each other. 
Their respective amino acid sequences on the right constitute a valuable signal that reflects their differences.}
    \label{fig:graph-vs-attributes}
\end{figure}

The use of entity attributes brings about the possibility of incorporating a wealth of multi-modal domain-specific data into the already extensive information present in the graph structure of the KG.
Crucially, attribute data can have a complementary relationship with the information in the graph: while two proteins might have the same relations with given molecules, their amino acid sequences might be different, as we illustrate in Fig. \ref{fig:graph-vs-attributes}.
Having access to attribute data can thus be fundamental for distinguishing entities that are identical according to the topology of the KG.

We are interested in knowledge graphs where entities can be associated with attributes of specific modalities (e.g., amino acid sequences for proteins), that encode important entity attributes.
Formally, we define such a KG as a tuple $G = (V, R, E, D, d)$, where $V$ is a set of nodes, $R$ a set of relations, $E$ a set of triples (or \emph{edges}) of the form $(h, r, t)$ with $h, t\in V$ and $r\in R$.
We refer to $h$ as the head, $r$ as the relation, and $t$ as the tail of the triple.
$D$ is a dataset containing attribute data of entities, and $d$ a partial function $d: V_d \rightarrow D$ where $V_d \subseteq V$ is the subset of entities with attribute data, and $d(v_i)$ retrieves the attribute data of entity $v_i$. 
We define $d$ as a partial function to formally support KGs for which attribute data is available for only a subset of the entities, which is a common scenario when dealing with large and heterogeneous graphs.

The information contained in KGs is often incomplete because certain facts are unknown or not considered important enough to make explicit. In other cases, changes in the domain could require the addition of new entities, new edges, or updates to existing ones.
This affects their ability to support the derivation of insights and scientific discovery.
Instead of retrieving existing relations between entities in the graph, a scientist might be interested in relations that are not yet present, but are nonetheless \emph{likely}~\cite{morselli2021network}.
These inferred relations could lead to the recommendation of a new hypothesis to be evaluated, such as an untested interaction between a protein and a molecule.

To tackle this scenario, different approaches have been proposed in machine learning literature that seek to exploit the patterns in the graph to train \emph{link prediction models}, which generalize to edges not observed during training.
In this work, we are concerned with the problem of link prediction over biomedical KGs with entity attributes.
Various challenges arise when optimizing link prediction models from both graph structure and entity attributes, especially in the biomedical domain.
The multi-modal nature of entity attribute data imposes a computational burden on models, further complicated by the fact that the coverage of attribute data is rarely perfect across the entire graph.
We propose a modular framework for learning knowledge graph embeddings that incorporates entity domain attributes in a link prediction setting that is subjected to these challenges. 

\subsection{Link Prediction with Knowledge Graph Embeddings}
Knowledge graph embeddings have become an increasingly popular tool for link prediction~\cite{nickel2016review,wang2017kgsurvey}. They consist of a mapping from the entities and relations in the graph, to vectors in a space where functions are used to model the likelihood of a triple.

More formally, KG embeddings can be defined in terms of an embedding function $e: E \cup R \rightarrow \mathcal{X}$ that maps an entity or relation in the graph to an element of an embedding space $\mathcal{X}$, plus a scoring function $f: \mathcal{X}^3 \rightarrow \mathbb{R}$ that given the embedding of the entities and relation in a triple, computes a score that indicates the likelihood of a triple.
Denoting the embedding of entities and relations in boldface (e.g. $e(h) = \mathbf{h}$), the embeddings are optimized so that the score $f(\mathbf{h}, \mathbf{r}, \mathbf{t})$ is high for a triple $(h, r, t)$ that is part of the KG, and low for a triple that is not.

A large part of the research carried out in the field of KG embeddings is concerned with designing suitable spaces $\mathcal{X}$ alongside embedding and scoring functions that are scalable and expressive enough to learn patterns that might occur in a KG.
In the simplest of cases, the embedding function is a lookup operation that for each entity and relation, returns an element of $\mathcal{X}$.
We call these methods \emph{look-up table embeddings}, where the table corresponds to a matrix of embeddings and a lookup or index that can be used to retrieve a specific row from it.
A well known example of such a method is TransE~\cite{bordes2013transe}, which maps entities and relations to vectors in a Euclidean space $\mathbb{R}^n$ of some dimension $n$.
For a triple $(h, r, t)$, TransE employs a translational model such that the sum of embeddings $\mathbf{h} + \mathbf{r}$ is close to $\mathbf{t}$.
This results in a scoring function that computes how close the vectors are:
\begin{equation}
    f_{\text{TransE}}(\mathbf{h}, \mathbf{r}, \mathbf{t}) = -\Vert\mathbf{h} + \mathbf{r} - \mathbf{t} \Vert.
\label{eq:transe-score}
\end{equation}

Other examples of lookup table embeddings are ComplEx~\cite{trouillon2016complex} and RotatE~\cite{sun2019rotate}, which instead use vectors defined over a complex-valued space $\mathbb{C}^n$, and scoring functions that perform multiplications or rotations over such vectors.
While these models have shown promising results in the task of predicting missing links between entities (also known as the \emph{link prediction} task), they suffer from two fundamental limitations.

First, they are optimized using only the structure of the graph, discarding other informative signals.
In particular, these representations ignore important information such as entity attributes that, as we outlined before, can contain critical information for differentiating between two entities in the graph.
Second, the mapping from the entities and relations to some embedding in the lookup table has to be fixed before running the training algorithm.
If new entities are added to the KG, the embeddings for these entities are not defined, which results in the inability to predict any links involving the new entities.
This limitation is at odds with the dynamic nature of KGs, especially in the biomedical domain.

Knowledge graph embedding methods that can compute predictions for entities not seen during training are known as \emph{inductive} link prediction models~\cite{xie2016dkrl,teru2020inductive,daza2021inductive,galkin2022nodepiece}.
An effective method to achieve this is to rely on entity attributes, which we will discuss next.

\subsection{Incorporating attributes in knowledge graph embeddings}

Instead of directly mapping entities in a KG to an embedding, some methods have proposed embedding functions that take as input the attributes of an entity~\cite{xie2017image,tay2017multitask,wang2018numericembedding,pouya2018embedding,kristiadi2019literale,xie2016dkrl,wang2021kepler,daza2021inductive,ektefaie2023multimodal}.
The embedding function can then be thought of as an \emph{encoder}, that can be designed through the use of neural network architectures to fit to specific modalities of the attribute data.
The task of the encoder is to process this data and output the embedding of the entity, which can in turn be passed to a particular scoring function.
During training, the encoder is optimized to map entity attributes to embeddings that are helpful for predicting links between entities.
Once training is finished, the encoder is able to produce embeddings for entities not seen during training as long as they are paired with their respective attribute data, thereby achieving the goal of inductive link prediction.

Recent methods that incorporate attributes assume that all entities in the graph are paired with a textual description, which is turned into a sequence of word embeddings and passed to a neural network that processes and aggregates the embeddings into a single vector~\cite{daza2021inductive,wang2022simkgc,markowitz2022statik,safavi2022cascader}.
Moreover, recent methods rely on pretrained language models such as BERT~\cite{devlin2019bert} as the backbone of the attribute encoder.
This allows the KG embeddings to benefit from the expressive architecture of such language models, and the encoded knowledge they contain through the vast amount of data used to pretrain them.

While previous methods allow encoding entity attributes into embeddings, they work on two fundamental assumptions: i) that all the attributes of entities share the same modality (e.g. text), and ii) that such information is available for all entities in the graph.
Biomedical KGs, especially those composed of data from different original sources are, on the contrary, characterized by heterogeneous entity types, each of which is associated with data of different modalities.
For example, disease entities can be described using text, molecules can be represented as small graphs of atoms and bonds, and proteins as sequences of amino acids.
It is therefore unclear how to directly transfer the results of encoder-based KG embeddings designed for text attributes, to the case of biomedical KGs where a textual representation might be natural only for some types of entities.
Furthermore, such methods do not provide a mechanism for dealing with entities with missing attribute data.

\subsection{Knowledge Graph Embedding Models in Biology.} 

KG embeddings can be used to tackle many LS problems, such as drug repurposing and target identification in the drug development process.
Previous works have shown that this can be achieved by collecting a sufficiently large knowledge graph that links entities such as proteins, diseases, genes, and drugs, and then training a model for various prediction tasks~\cite{ali2019biokeen,nelson2019embed,walsh2020biokg,mohamed2020trimodel,alshahrani2021application,ye2022tensorfac,gema2023knowledge}.
In particular, some of these works transfer the findings from the literature of KG embeddings--especially lookup table embeddings--to the biomedical domain, finding that the embeddings are effective at solving problems such as Drug-Target Interaction (DTI) prediction~\cite{karim2019drug,alshahrani2021application,ye2022tensorfac}.

While these results are promising, prior work has also highlighted challenges for using KG embeddings in practice~\cite{walsh2020biokg,ali2019biokeen}.
In particular, KG embeddings perform better on well-studied biomedical entities, because more detailed information is available about them that can be incorporated in the graph.
This can cause low link prediction performance on understudied entities~\cite{rossi2020knowledge}.
Moreover, by using lookup table embeddings, these methods are still limited to the structural information contained in the graph, and do not allow for predictions involving new entities~\cite{alshahrani2021application}.

Recent work has investigated the effect of incorporating entity attributes into embeddings for various applications in the biomedical domain~\cite{choi2021identifying,alshahrani2022combining,ren2022biomedical,su2022attention,zhang2022mkge,zhu2022multimodal}.
The proposed methods are specialized to certain data modalities, such as textual descriptions~\cite{choi2021identifying,alshahrani2022combining,zhu2022multimodal}, or SMILES molecular representations~\cite{ren2022biomedical,su2022attention,zhang2022mkge}.
These methods are designed for biomedical KGs where the {\em data modality is the same for all entities} in the graph, and they also assume that attribute data is available for all entities.

Furthermore, some works have found that incorporating attribute data can outperform baseline methods based on the structure of the graph only~\cite{zhang2022mkge,su2022attention,ren2022biomedical,alshahrani2022combining}.
However, these works do not disclose whether or not hyperparameter optimization of baselines was carried out~\cite{alshahrani2022combining,ren2022biomedical,su2022attention}, or indicate that it was omitted~\cite{zhang2022mkge}.
We argue that adequate hyperparameter optimization is crucial for understanding the effect of incorporating attributes in comparison with methods based on the graph structure only, given that the latter can be quite sensitive to hyperparameters~\cite{ruffinelli2020dog,ali2022bringing,bonner2022understanding}.

\subsection{Research objective}\label{subsec:research_obj}

We set out to address the limitations in the literature by proposing a modular framework for learning embeddings in a KG with entity attributes.
Unlike previous work, we propose encoders for attributes of different modalities, while also allowing learning embeddings for entities with missing attributes.
Furthermore, we propose an efficient strategy for reducing the training runtime of computationally demanding attribute encoders.

To investigate the impact of incorporating attributes in our framework, we perform extensive hyperparameter optimization studies over baselines that do not consider entity attributes.
To gauge the performance of the resulting embeddings, we carry out a comprehensive evaluation on the tasks of link prediction and drug-protein interaction prediction.
In further analyses, we study which entity attribute encoders can address limitations of conventional KG embedding models, primarily in the ability to predict over low-degree entities in sparser regions of the graph.

\section{Methods}
In the following sections, we describe the biomedical KG used in our research, the proposed framework for learning embeddings, and the evaluation protocols used to assess its performance.

\subsection{The BioKG knowledge graph}

BioKG is a Knowledge Graph built for the purpose of supporting relational learning in complex biological systems~\cite{walsh2020biokg}, with a focus on protein and molecules (interchangeably referred to as drugs).
It was created in response to the problem of data integration and linking between various open biomedical knowledge bases, resulting in a compilation of curated data from 13 different biomedical databases, including DrugBank~\cite{knox2010drugbank}, UniProt~\cite{uniprot2015uniprot}, and MeSH~\cite{lipscomb2000medical}.

The pipeline for creating BioKG employs a module to programmatically retrieve, clean and link the different entities and relations found in different databases, by reconciling their identifiers.
This results in a biomedical KG containing over 2 million triples, where each entity has a uniform identifier linking it to publicly available databases. 
The coverage over multiple biomedical databases in BioKG presents a significant potential for training machine learning models on KGs with attribute data.
Additionally, the availability of identifiers in BioKG facilitates the search for entity attributes of different modalities.
These characteristics make BioKG an ideal choice for furthering research in the direction of machine learning on knowledge graphs with attribute data.

\begin{figure}[t]
    \centering
    \includegraphics[width=0.6\linewidth]{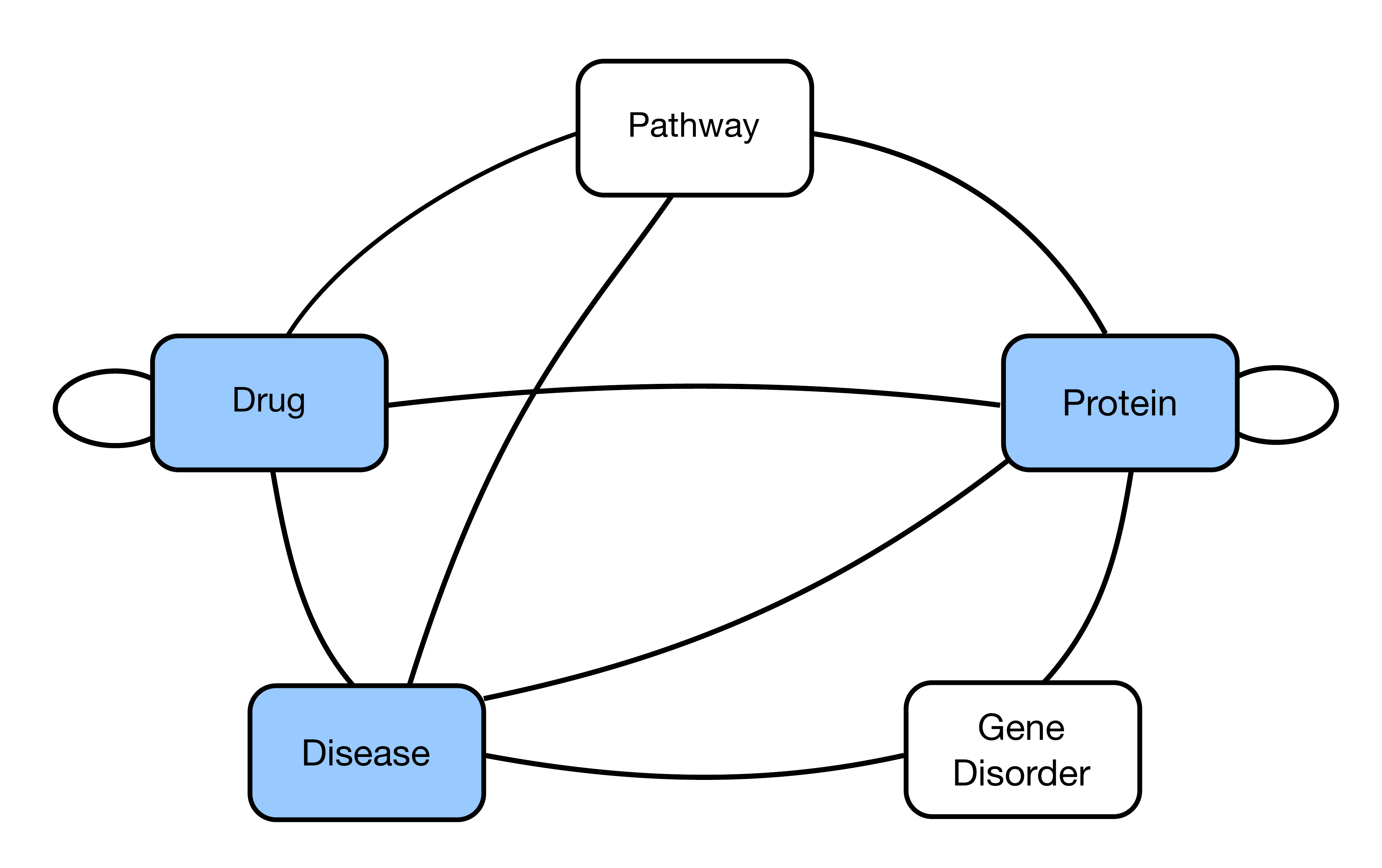}
    \caption{Visualization of the entity types in the BioKG graph. Edges between nodes denote the existence of any relation type. Shaded nodes indicate entities for which we retrieve attribute data.}
    \label{fig:biokg-schema}
\end{figure}

BioKG contains triples involving entities of 5 types: proteins, drugs, diseases, pathways, and gene disorders.
We use the publicly available repository~\footnote{Available at \url{https://github.com/dsi-bdi/biokg}} to compile BioKG from the original sources, as available on May 2022.
A visualization of the schema is shown in Fig.~\ref{fig:biokg-schema}, and the statistics of the graph are listed in Table~\ref{tab:biokg-stats}.

\begin{table}[t]
\caption{Statistics of the BioKG graph used in our work. 
The Attribute Coverage refers to the number of entities that have associated attribute data, which we retrieve for proteins, molecules, and diseases.}
\label{tab:biokg-stats}
\begin{tabular}{lrr}
\toprule
\textbf{Statistic}         & \textbf{Count} & \textbf{Attribute Coverage}\\
\midrule
Edges                      & 2,074,346      &  ---                       \\
Relations                  & 17             &  ---                       \\
Entity types               & 5              &  ---                       \\
\midrule
Entities                   & 106,337        & 73,131                     \\
\quad Proteins             & 59,664         & 59,551                     \\
\quad Drugs                & 8,808          & 7,766                      \\ 
\quad Diseases             & 5,814          & 5,814                      \\
\quad Other                & 32,501         & 0 \\
\bottomrule
\end{tabular}
\centering
\end{table}

Following our aim of learning embeddings that incorporate attribute data of different modalities, while taking into account entities with missing attribute data, we retrieve attributes for proteins, drugs, and diseases.
To do so, we take advantage of the uniform entity identifiers present in BioKG, that allow us to query publicly available API endpoints.
In particular, we retrieve amino acid sequences from Uniprot\footnote{\url{https://www.uniprot.org/}}, SMILES representations for drugs from DrugBank\footnote{\url{https://go.drugbank.com/}}, and textual descriptions of diseases from MeSH\footnote{\url{https://www.nlm.nih.gov/mesh/}}.
In a few cases where a textual MeSH scope note was unavailable for a disease, we use the text label for the disease as is, as this can contain useful information about the class, or anatomical categorization of a disease.
We report the resulting coverage of attribute data for the entity types of our interest in Table~\ref{tab:biokg-stats}.
The result is a KG containing attribute data of three modalities for approximately 68\% of the entities, while for the rest of the entities attribute data is missing, which are the desired characteristics of the KG for our research objective.

\subsection{Biomedical benchmarks}

One aspect that separates BioKG from other related efforts~\cite{williams2012open}, is the inclusion of benchmark datasets around domain-specific tasks such as Drug-Protein Interaction and Drug-Drug Interaction prediction.
This is crucial, as it allows for experimentation and comparison to standard benchmark datasets and approaches.
The benchmarks consist of pairs of entities for which a particular relationship holds, such as an interaction between drugs.
We use them for comparing the predictive performance of KG embeddings obtained when using the graph structure alone, and when incorporating attribute data.
In total, five benchmarks are provided with BioKG, all  considering different drug-drug and drug-protein interactions.

We focus on the DPI-FDA benchmark, which consists of drug target protein interactions of FDA approved drugs compiled from KEGG and DrugBank databases. 
It is comprised of 18,928 drug-protein interactions of 2,277 drugs and 22,654 proteins.
This benchmark is an extension of the DrugBank\_FDA~\cite{wishart2008drugbank} and Yamanishi09~\cite{yamanishi2008prediction} datasets which have 9,881 and 5,127 DPIs respectively.
We refer to it as the DPI benchmark.

\subsection{Data partitioning}
In order to train and evaluate different KG embedding methods, we need to specify sets of triples for training, validation, and testing.
Considering that the original BioKG dataset does not provide predefined splits, we generate our own.
Since we aim to train multiple embedding methods on BioKG, and evaluate their performance over the benchmarks, we must guarantee that the KG and the benchmarks are \emph{decoupled}, so that there is no data leakage from the benchmarks into the set of KG triples used for training.

Our strategy for partitioning the data is as follows:
\begin{enumerate}
    \item We merge the five benchmarks provided in BioKG into a single collection of entity pairs.
    \item For every pair of entities $(x, y)$ in the collection, we remove all triples from BioKG in which $x$ and $y$ appear together, regardless of which one is at the head or tail of the triple.
    \item Removing triples from BioKG in step 2 might lead to the removal of an entity entirely from the graph.
    \item We split the resulting triples in BioKG into training, validation, and test sets, ensuring that all entities occur at least once in the training set of triples.
\end{enumerate}

The last two steps are a requirement for lookup table embeddings.
These methods require that all entities are observed in the training set of triples, so that we can obtain embeddings for evaluating i) link prediction performance on the validation and test triples, and ii) prediction accuracy on the DPI benchmark.

The proportions used for selecting training, validation, and test triples are 0.8, 0.1, and 0.1, respectively.
We present statistics of the generated splits in Table~\ref{tab:splits-stats}.
In the DPI benchmark, we end up with 18,678 pairs that we used to train classifiers and evaluate them via cross-validation.

To further research in this area, we make all the data we collected publicly available, comprising the attribute data for proteins, molecules, and diseases, and the splits generated for training, validation, and testing of machine learning models trained on BioKG\footnote{Available at \url{https://doi.org/10.5281/zenodo.8005711}.}.

\begin{table}[t]
\caption{Statistics of the BioKG splits of triples used for training, validation, and testing, after removing triples overlapping with the benchmarks.}
\label{tab:splits-stats}
\begin{tabular}{lrrcc}
\toprule
                   & \textbf{Total} & \textbf{Training} & \textbf{Validation} & \textbf{Test} \\
\midrule                            
BioKG              &     1,852,262  &        1,481,809  &             185,226 &       185,227  \\
\bottomrule
\end{tabular}
\centering
\end{table}

\subsection{BioBLP}
We now detail our new method, BioBLP, proposed as a solution to the problem of learning embeddings of entities in a KG with attributes.
We design it as a extension of BLP~\cite{daza2021inductive} to the biomedical domain that incorporates the three data modalities available to us in the BioKG graph.
BLP is a method for embedding entities in KGs where all entities have an associated textual description.
At its core lies an encoder of textual descriptions based on BERT, a language model pretrained on large amounts of text~\cite{devlin2019bert}.
In this work, we propose a more general formulation that considers multiple modalities as well as missing attribute data.
By supporting KGs that contain entities \emph{with} and \emph{without} attributes, we can exploit the full extent of the information contained in the graph, rather than limiting the data available for training to one of the two cases.

BioBLP is a model for learning embeddings of entities and relations in a KG with attribute data of different modalities.
It is comprised by four components: a set of attribute encoders, lookup table embeddings for entities with missing attributes and for relation types, a scoring function, and a loss function.

\paragraph{Attribute encoders}
Let $G = (V, R, E, D, d)$ be a KG with attribute data.
BioBLP contains a set of $k$ modality specific \emph{encoders} $\lbrace e_1, \ldots, e_k\rbrace$.
Each of the $e_i$ are functions of the attributes of an entity that output a fixed-length embedding.
If we denote as $D_i\subset D$ the subset of attribute data of modality $i$ (e.g. the subset of protein sequences), then formally each encoder is a mapping $e_i: D_i\rightarrow \mathcal{X}$, where $\mathcal{X}$ is the embedding space shared by all entities. 
We design the encoders as neural networks with learnable parameters, whose architecture is tailored to the input modality.

\paragraph{Lookup table embeddings}
For entities that are missing attribute data, we employ a lookup table of embeddings.
For each entity in $V$ with missing attributes, we assign it an initial embedding by randomly sampling a vector from $\mathcal{X}$.
We follow the same procedure for all relations in $R$.
These embeddings are then optimized, together with the parameters of the attribute encoders in $F$.
In practical terms, this means that when attributes are not available at all, BioBLP reduces to lookup table embedding methods such as TransE~\cite{bordes2013transe}.
We implement the lookup table embeddings via matrices $\mathbf{W}_e$ and $\mathbf{W}_r$ for entities and relations, respectively, with each of the rows containing a vector in $\mathcal{X}$.
The embedding function of an entity without attributes is an additional encoder $e_0$ such that $e_0(h)$ returns the row corresponding to entity $h$ in $\mathbf{W}_e$.
Similarly for relations, we define its embedding via a function $e_r$ the retrieves the embedding of a relation from the rows of $\mathbf{W}_r$.

\paragraph{Scoring function}
Once the attribute encoders and the lookup table embeddings have been defined, for a given triple $(h, r, t)$ we can obtain the corresponding embeddings $\mathbf{h}, \mathbf{r}, \mathbf{t}\in\mathcal{X}$.
These can be passed to any scoring function for triples $f: \mathcal{X}^3\rightarrow \mathbb{R}$, such as the TransE scoring function defined in Eq.~\ref{eq:transe-score}.
The scoring function also determines the type of embedding space $\mathcal{X}$.
For TransE, this corresponds to the space of real vectors $\mathbb{R}^n$ while for ComplEx and RotatE it is the space of complex vectors $\mathbb{C}^n$.

\paragraph{Loss function}
The loss function is designed so that it can be minimized by assigning high scores to observed triples, and low scores to corrupted triples~\cite{ali2022bringing}.
Corrupted triples are commonly obtained by replacing the head or tail entities in a triple $(h, r, t)$ by an entity sampled at random from the KG.
For such a triple, the loss function can thus be defined as a function $\mathcal{L}(f(\mathbf{h}, \mathbf{r}, \mathbf{t}), f(\tilde{\mathbf{h}}, \mathbf{r}, \tilde{\mathbf{t}}))$.
An example is the margin ranking loss function:
\begin{equation}
    \mathcal{L}(f(\mathbf{h}, \mathbf{r}, \mathbf{t}), f(\tilde{\mathbf{h}}, \mathbf{r}, \tilde{\mathbf{t}})) = \max(0, f(\tilde{\mathbf{h}}, \mathbf{r}, \tilde{\mathbf{t}})- f(\mathbf{h}, \mathbf{r}, \mathbf{t}) + m),
\end{equation}
which forces the score of observed triples to be higher than for corrupted triples by a margin of $m$ or more.

We optimize the parameters of the attribute encoders and the lookup table embeddings via gradient descent on the loss function.
If we denote by $\boldsymbol{\theta}$ the collection of parameters in the entity encoders plus the lookup table embeddings, the training update rule during for a specific triple is the following:
\begin{equation}
    \boldsymbol{\theta}^{(t + 1)} = \boldsymbol{\theta}^{(t)} - \alpha \nabla_\theta \mathcal{L}(f(\mathbf{h}, \mathbf{r}, \mathbf{t}), f(\tilde{\mathbf{h}}, \mathbf{r}, \tilde{\mathbf{t}})),
\end{equation}
where $t$ is a training step, $\nabla_\theta$ is the gradient with respect to $\boldsymbol{\theta}$ and $\alpha$ is the learning rate.
We outline the algorithm for training BioBLP in Algorithm~\ref{alg:bioblp}.

\begin{algorithm}[t]
\KwIn{Knowledge graph $(V, R, E, D)$ with entity-to-attribute map $d: V\rightarrow D$ \newline
    Modality mapping $\texttt{modality\_of}: D \rightarrow \lbrace 0, 1, \ldots, k\rbrace$ \newline
    Embedding functions of entities and relations: $\lbrace e_0, \ldots, e_k\rbrace, e_r$ \newline
    Scoring function $f$\newline
    Loss function $\mathcal{L}$\newline
    Number of steps $T$, Learning rate $\alpha$
    }
    \SetKwFunction{fembed}{emb}
    \SetKwProg{Fn}{Function}{:}{}
    \Fn{\fembed{entity $x$}}{
        $m$ $\gets$ \texttt{modality\_of}$(d(x))$\;
        \uIf{m = 0}{
            $\mathbf{x} = e_0(x)$ \;
        }
        \Else{
            $\mathbf{x} = e_m(d(x))$ \;
        }
        \KwRet $\mathbf{x}$\;
    }
    $\boldsymbol{\theta}^{(0)} \gets$ parameters in $\lbrace e_0, \ldots, e_k\rbrace, e_r$\;
    \For{$t = 1\ldots T$} {
        \For{$(h, r, t) \in E$}{
            $(\tilde{h}, r, \tilde{t}) \gets$ corrupt $(h, r, t)$\;            
            $\boldsymbol{\theta}^{(t)}\gets \boldsymbol{\theta}^{(t-1)} - \alpha\nabla_\theta\mathcal{L}(f(\texttt{emb}(h), e_r(r), \texttt{emb}(t)), f(\texttt{emb}(\tilde{h}), e_r(r), \texttt{emb}(\tilde{t}))$
        }
    }
 \Return{$\boldsymbol{\theta}^{(T)}$}
 \caption{BioBLP: Learning embeddings in a KG with entity attributes.}
 \label{alg:bioblp}
\end{algorithm}

\subsubsection{Implementing BioBLP}

The modular formulation of BioBLP is general enough to enable experiments with different choices for each of the modules, and presents an opportunity for incorporating domain knowledge into the design of different attribute encoders.
In this section we describe the particular choices that we make for training BioBLP with the BioKG graph.

Considering that we retrieve attribute data for proteins, drugs, and diseases in BioKG, we define encoders for these three types.
Recent works have demonstrated the effect of using pretrained embeddings in link prediction models that use textual descriptions~\cite{daza2021inductive,wang2022simkgc,markowitz2022statik}.
Pretraining is done on a task for which a large dataset is available, such as masked language modeling~\cite{devlin2019bert}.
Intuitively, the use of large pretrained models provides a better starting point for training, compared to randomly initializing an encoder and training for link prediction.
In the context of the biomedical domain, we explore this strategy by considering three pretrained models as attribute encoders: ProtTrans~\cite{elnaggar2021prottrans}, MolTrans~\cite{morris2020transformer}, and BioBERT~\cite{jinhyuk2019biobert}.

\paragraph{Protein Encoder}
ProtTrans is a model for encoding amino acid sequences of proteins, where each of the input tokens is an amino acid~\cite{elnaggar2021prottrans}.
The output is a matrix of ``contextualized'' amino acid embeddings of size $n_e\times l$, where $l$ is the length of the sequence and $n_e$ the dimension of the ProtTrans embeddings. 
Encoding proteins with ProtTrans in BioKG is challenging: we found that some sequences can have lengths up to 35,213, which greatly increases memory requirements.
To reduce memory usage during training, we choose not to optimize the parameters of the ProtTrans encoder.
Given the output matrix of amino acid embeddings, we obtain a single protein embedding by taking the average over the sequence, and then linearly projecting the result to the embedding space of dimension $n$ via a matrix of size $n\times n_l$ with learned parameters.

\paragraph{Molecule Encoder}
MolTrans is a model initially trained to learn representations of molecules, by translating them from their SMILES representation to their IUPAC name~\cite{morris2020transformer}.
We use it to obtain embeddings for molecules in BioKG for which we have a SMILES representation.
The output of MolTrans is a matrix of embeddings of size $n_e\times l$, where $l$ is the length of the SMILES string.
Similar to the case of the protein encoder, we do not optimize the parameters of MolTrans due to memory constraints.
Instead, we apply a layer of self-attention~\cite{vaswani2017attention} and select the output embedding at the first position, which corresponds to a special marker for the beginning of the sequence.
This embedding is then linearly projected to an embedding of size $n$.
In the case of the molecule encoder, the learnable parameters are the weights in the self-attention layer, and the linear projection matrix.

\paragraph{Disease Encoder}
We rely on BioBERT~\cite{jinhyuk2019biobert} to encode textual descriptions of entities.
BioBERT is a language model trained to predict words that have been removed from some input text, which is also known as the masked language modeling task~\cite{devlin2019bert}.
In contrast to models like BERT~\cite{devlin2019bert} that have been trained over diverse sources from the internet, BioBERT has been pretrained with scientific articles obtained from PubMed and PMC.
Given an input textual description, we select the output embedding of the special marker at the start of the description, and then project it to an embedding of size $n$.
As opposed to the protein and molecule encoders, we optimize all the layers in the BioBERT encoder as well as the projection layer.
We can afford this by limiting the length of the descriptions to a maximum of 64 tokens, which helps to curb the memory requirements during training.
Previous work has shown that further increasing the length of text encoders for embedding entities in KGs brings dimishing returns~\cite{daza2021inductive}.

\begin{figure}[t]
    \centering
    \includegraphics[width=0.95\linewidth]{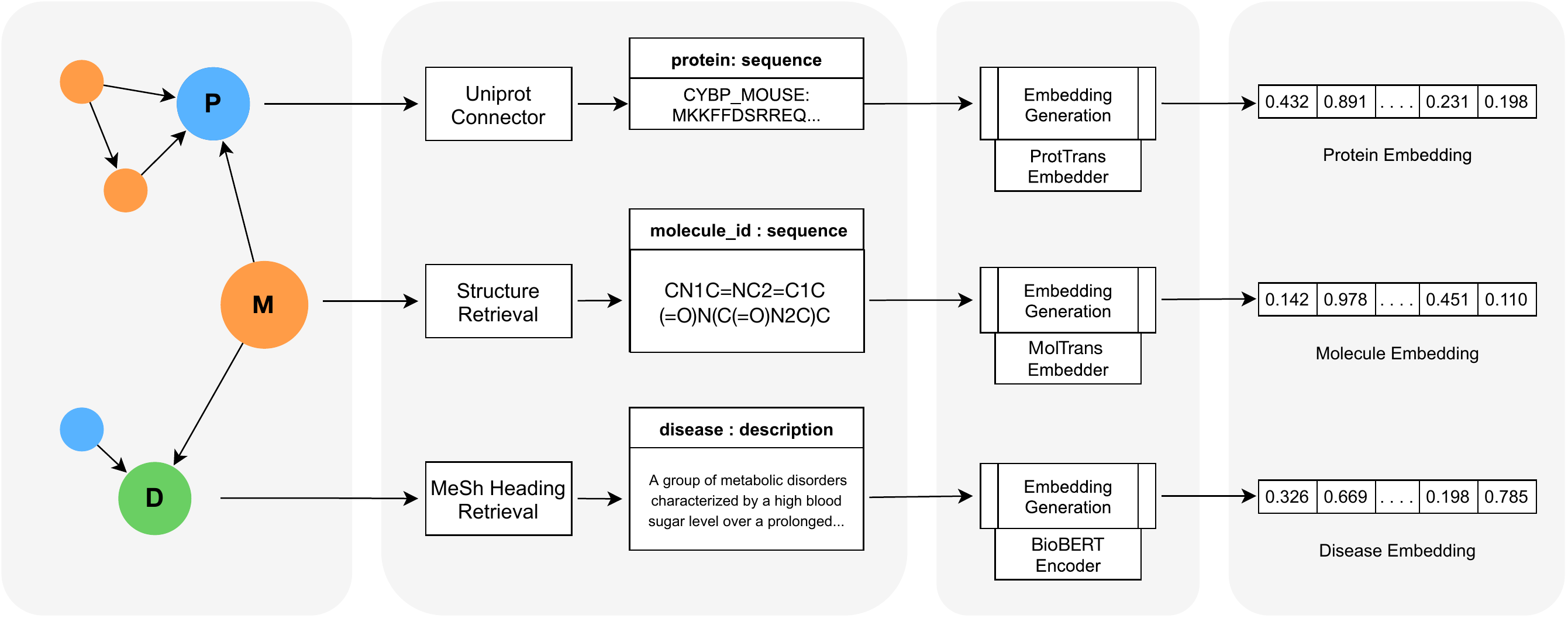}
    \caption{The architecture of BioBLP for the BioKG graph. On the left we illustrate the graph with the three entity types that have attribute data: proteins (P), molecules (M, representing drugs), and diseases (D). We omit entity types without attributes for clarity. For the three types, we retrieve relevant attribute data, which are the inputs to encoders for each modality. The output of the encoders is a vector representation of the entities.}
    \label{fig:bioblp}
\end{figure}

In summary, we provide an implementation of BioBLP for the BioKG graph that uses encoders of amino acid sequences for proteins, SMILES representations for molecules, and textual descriptions of diseases.
Following the BioBLP framework, other entity types like pathways and gene disorders are embedded using lookup table embeddings.
We illustrate the architecture in Fig.~\ref{fig:bioblp}.

Previous research has shown that in spite of continuous developments in the area of KG embeddings, the choice of scoring function is a matter of experimentation and the results are highly data-dependent~\cite{ruffinelli2020dog,daza2021inductive,ali2022bringing}.
For these reasons, we experiment with three scoring functions: TransE~\cite{bordes2013transe}, ComplEx~\cite{trouillon2016complex}, and RotatE~\cite{sun2019rotate}.
For the loss functions, we consider the margin ranking loss, the binary cross-entropy loss, and the cross-entropy loss~\cite{ali2022bringing}.
Our implementation uses the PyKEEN library~\cite{ali2021pykeen} and is publicly available\footnote{Available at \url{https://github.com/elsevier-AI-Lab/BioBLP}.}.

\subsubsection{Efficient model pretraining}
Training KG embedding methods that use encoders of attribute data has been shown to significantly increase the required computational resources for training~\cite{daza2021inductive,markowitz2022statik,wang2022simkgc}.
Furthermore, BioBLP consists of a set of attribute encoders, together with lookup table embeddings for entities with missing attributes.
Training such a model from scratch requires all the encoders to adapt together for the link prediction objective.
If lookup table embeddings are initialized randomly as is customary, attribute encoders will adjust to their random-level performance, greatly slowing down the rate of learning.
We thus propose the following strategy to overcome this problem:
\begin{enumerate}
    \item Train a BioBLP model with lookup table embeddings for \emph{all} the entities in the graph.
    \item Train a BioBLP model with encoders for entities with attribute data, and for entities without such data, initialize their lookup table embeddings from the model trained in step 1.
\end{enumerate}

Step 1 is much cheaper to run since it does not make use of any attribute encoders.
This speeds up training, and facilitates hyperparameter optimization.
In step 2, the encoders no longer have to adapt to randomly initialized lookup table embeddings, but to pretrained embeddings that already contain useful information about the structure of the graph, which reduces the number of training steps required.

\subsection{Evaluation}
\label{sec:evaluation}

In our experiments, we aim to answer the following research question: how do the different attribute encoders in BioBLP compare with traditional KG embedding methods that rely on the structure of the graph only?
We focus on two tasks to answer this question: link prediction, and the DPI benchmark present in BioKG.

\subsubsection{Link prediction}
The link prediction evaluation is standard in the literature of KG embeddings~\cite{ruffinelli2020dog,ali2022bringing}.
Once the model is trained, the evaluation is carried out over a held-out test set of triples that is disjoint from the set of triples used during training.

For a triple $(h, r, t)$ in the test set, a list of scores for \emph{tail prediction} is obtained by fixing $h$ and $r$ and computing the scoring function $f(\mathbf{h}, \mathbf{r}, \mathbf{x})$, for all possible entity embeddings $\mathbf{x}$.
The list of scores is sorted from high to low, and the rank of the score for the true tail entity $t$ is determined.
We repeat this for all triples in the test set, collecting a list of ranks for tail prediction.
For head prediction, the process is similar, but instead we fix $r$ and $t$ and collect the ranks of the true head entity $h$ given scores for the head position.
We collect the ranks for head and tail prediction in a list $(p_1, \ldots, p_m)$.
From this list we compute the Mean Reciprocal Rank, defined as
\begin{equation}
    \text{MRR} = \frac{1}{m}\sum_{i = 1}^m \frac{1}{p_i},
\end{equation}
and Hits at k, defined as
\begin{equation}
    \text{H@k} = \frac{1}{m} \sum_{i = 1}^m \mathbbm{1}[p_i \leq k],
\end{equation}
where $\mathbbm{1}[\cdot]$ is an indicator function equal to 1 if the argument is true and zero otherwise.
It is important to note that by averaging over the set of evaluation triples, these metrics are more influenced by high-degree nodes, simply because they have more incident edges.

We use the MRR and H@k metrics to report the performance of the embeddings obtained with BioBLP.
As baselines that do not incorporate attribute data, we consider TransE, ComplEx, and RotatE.
For a direct comparison, we experiment with variants of BioBLP that use the same scoring functions as in these baselines.
This allows us to determine when improvements are due to the choice of scoring function, rather than a result of incorporating attribute data.

\paragraph{Hyperparameter Study}
Previous research has shown that KG embedding models and the optimization algorithms used to train them can be sensitive to particular choices of hyperparamenters~\cite{ruffinelli2020dog}, including in the biomedical domain~\cite{bonner2022understanding}.
Hence, a key step in determining the performance of our baselines against models that incorporate attributes is an exhaustive optimization of their hyperparameters.
For each of the three baselines we consider, we carry out a hyperparameter study with the aim of optimizing their performance before comparing them with BioBLP.
We focus on four hyperparameters as listed in Table~\ref{tab:hparam-values}, where we also describe the intervals and values we considered.
For each baseline, we used Bayesian optimization to search for hyperparameters, running a maximum of 5 trials in parallel.
We ran a total of 90 different hyperparameter configurations with TransE, 180 with ComplEx, and 180 with RotatE.
For TransE we only consider the margin ranking loss function, while for ComplEx and TransE we experiment with the binary cross-entropy and cross-entropy, which leads to twice the number of experiments.

\begin{table}[t]
    \caption{Intervals and values used in the hyperparameter study we carried out to optimize the TransE, ComplEx, and RotatE baselines.}
    \label{tab:hparam-values}
    \begin{tabular}{ll}
    \toprule
    Hyperparameter        & Values \\
    \midrule
    Learning rate         & Log-uniformly sampled from $[1\times 10^{-3}, 1.0]$      \\
    Regularization weight & Log-uniformly sampled from $[1\times 10^{-6}, 1\times 10^{-3}]$ \\
    Batch size            & $\lbrace128, 256, 512, 1024\rbrace$ \\
    Loss function         & Margin ranking, binary cross-entropy, cross-entropy
    \end{tabular}
\end{table}

\subsubsection{Benchmarks}
\label{benchmarks}

We evaluate the utility of entity embeddings by testing their performance as input features for classifiers trained on related biomedical tasks.
We focus on the DPI benchmark obtained from BioKG.
As it is common in literature, we interpret this task as binary classification~\cite{nascimento2016kronrls, hao2017blmnii, olayan2018ddr}.
We use the pairs in the benchmarks as positive instances to train the classifiers.
A common issue in modelling efforts in biomedical domain is the lack of true negative samples in public benchmark datasets. 
Therefore, we sample negatives from the combinatorial space assuming a pair that is unobserved in the dataset is a candidate negative sample~\cite{nascimento2016kronrls, hao2017blmnii, olayan2018ddr, mohamed2020trimodel}. 
Furthermore, we filter out any triples appearing in BioKG from the list of candidate negatives.
We use a ratio of positive to negatives of 1:10.

For a given pair of entities, we concatenate their embeddings into a single vector that is used as input to a classifier.
We evaluate a number of baselines for computing embeddings, including random embeddings, entity embeddings that do not rely on the graph, and KG embeddings.
We compare these to embeddings obtained with BioBLP.
In particular, we use the following embeddings:
\begin{itemize}
    \item \textbf{Random}. Every entity is assigned a randomly generated feature vector from a normal distribution. We use this method to determine a lower bound of chance performance. Each entity is assigned an 128-dimensional embedding.
    \item \textbf{Structural}. Drug entities are represented by their MolTrans embeddings of 512 dimensions, and proteins are represented by their ProtTrans embedding, averaged across tokens resulting in a 1024-dimensional vector.
    \item \textbf{KG}. Entities are represented by TransE, ComplEx, and RotatE embeddings trained on BioKG. The entity embeddings from these models all have 512 dimensions.
    \item \textbf{BioBLP}. Entities are represented from the attribute encoders in BioBLP models trained using the RotatE scoring function, on the BioKG graph. The entity embeddings from the BioBLP-D, BioBLP-M and BioBLP-P models have 512 dimensions.
\end{itemize}

When using Structural Embedding, it may occur that a particular entity does not have a corresponding feature, for example, because we are missing the chemical structure embeddings for a set of molecules (as detailed in Table~\ref{tab:biokg-stats}) because these are biologic therapies and do not have a SMILES property.
In these cases, we impute the missing embedding by the mean of all entity embeddings for that entity type. 

\paragraph{Training and Evaluation}
We experiment with Logistic Regression (LR), Random Forest (RF) and Multi Layer Perceptron (MLP) models for classification.
To aid the training of the classifiers in the imbalanced data setting, we use class weights inversely proportional to the class frequency in the LR and RF models, and undersample the majority class in the training of the MLP models.

The performance estimates of the classification models is based on 5-fold stratified cross-validation.
The folds are kept consistent for training and testing across all models.
For each fold, hyperparameter optimization is performed to establish optimal model parameters on the training split.
Ten percent of the training split is reserved for validation in the tuning procedure.
We use Optuna to perform the hyperparameter search, which uses a tree-structured Parzen estimator algorithm for sampling the space of hyperparameters.
For all models, 10 trials are performed to find the best parameters on the validation set that maximize performance on the area under precision-recall curve (AUPRC) and area under ROC curve (AUROC) metrics.
From these trials the best parameters are determined and the model is retrained on the full training set and scored against the test set.

\paragraph{Metrics}
In the evaluation procedure of the classification models we use Precision, Recall, F1, AUROC and AUPRC values. 
The AUROC is commonly reported in bioinformatics literature but is argued to be an overly optimistic measure of performance~\cite{takaya2015}. 
For this reason, we additionally report the AUPRC because this metric is particularly informative in imbalanced datasets due to its sensitivity to true positives among all positive predictions. 

\section{Results}

In this section we describe our findings on the link prediction performance of embedding methods that do not use attributes, and the effect of incorporating them with BioBLP.
We additionally present the results on the DPI benchmark, and we then provide a fine-grained analysis of the performance provided by BioBLP.

\subsection{Hyperparameter study}

\begin{figure}[t]
    \centering
    \includegraphics[width=0.9\linewidth]{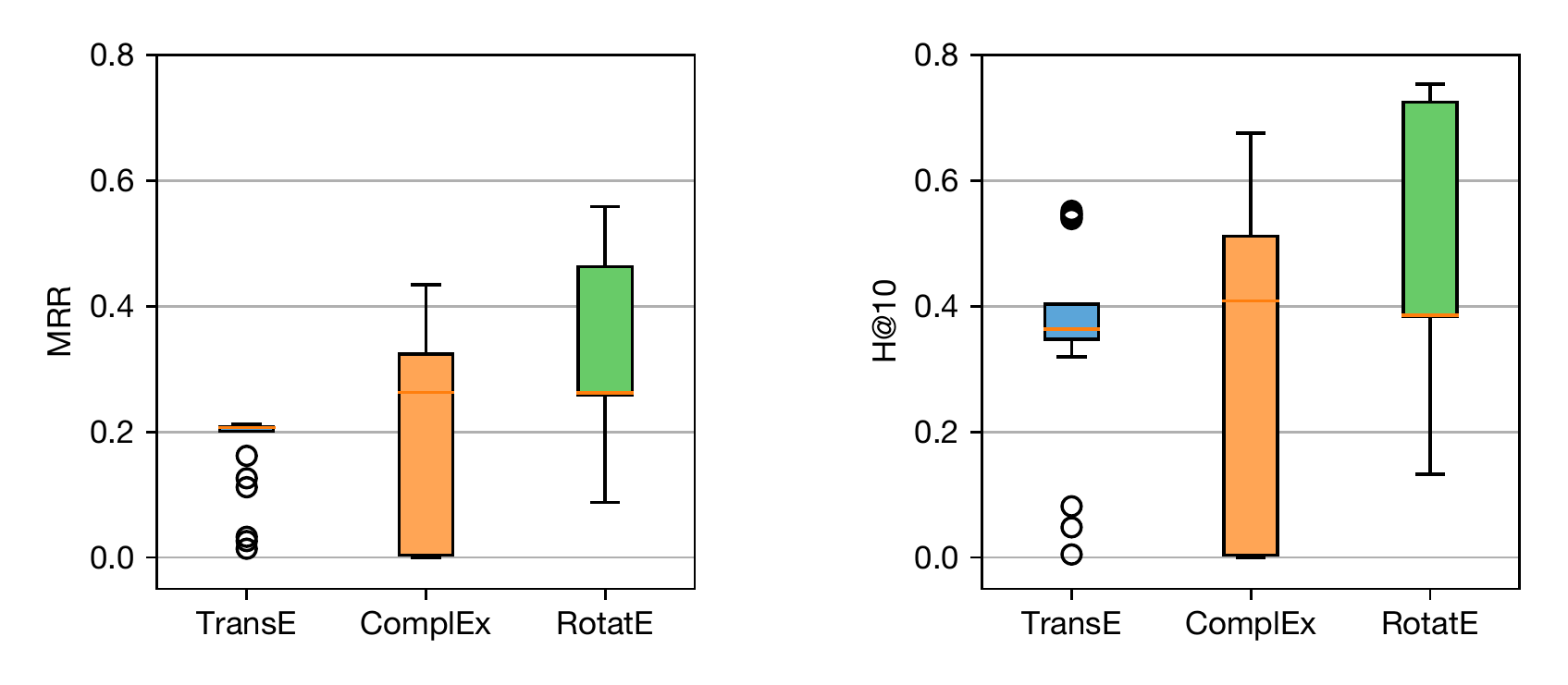}
    \caption{Distribution of MRR (left) and H@10 (right) performance metrics for link prediction, across all values explored during hyperparameter optimization of the baseline models. TransE performs rather consistently, but poorly overall. ComplEx results on a broad range of outcomes, indicating that it is very sensitive to hyper-parameters. RotateE performs similar to ComplEx when comparing their medians, but has a smaller spread and with specific parameters it results in the best performance.
}
    \label{fig:hparam-opt-results}
\end{figure}

We illustrate in Fig.~\ref{fig:hparam-opt-results} the distribution of link prediction metrics for the three baselines that we consider, across all the configurations explored during the hyperparameter study, evaluated on the test set of triples.
These results confirm the importance of thoroughly exploring the space of hyperparameters, as the three baselines exhibit high variance in the final performance.
For example, we observe that for specific configurations, TransE can outperform RotatE, even though by design RotatE is more expressive at modeling different kinds of relations~\cite{sun2019rotate}.

Across all the hyperparameter trials that we run, we observe that RotatE yields consistently better results, with the worse RotatE configuration outperforming multiple configurations of TransE and RotatE.
We select the best configurations of these three models based on the validation set performance, and we use them in the next set of experiments.

\subsection{Link prediction performance}

We present link prediction results of BioBLP under different configurations, corresponding to variants along three axes: the scoring function, the type of encoder, and whether or not the model includes our proposed pretraining strategy.
For scoring functions, we experiment with TransE, ComplEx, and RotatE, and for encoders, we experiment with Proteins (P), Molecules (M), and Diseases (D).
An example of a specific configuration for a BioBLP model that uses the TransE scoring function and implements a molecule encoder is TransE+BioBLP-M.

\begin{table}[t]
    \caption{Link prediction performance on the BioKG dataset (in percent).}
    \label{tab:link-prediction-results}
    \begin{tabular}{lcrrrr}
        \toprule          
        Method            & Pretrained &    MRR   &     H@1  &     H@3  &     H@10 \\
        \midrule
        TransE            &            &    20.61 &    12.15 &    22.14 &    36.27 \\
        \quad + BioBLP-P  &            &    14.76 &     8.42 &    15.60 &    25.92 \\
        \quad + BioBLP-P  & \checkmark &    14.87 &     8.47 &    15.67 &    26.16 \\
        \quad + BioBLP-M  &            &     6.56 &     4.46 &     6.78 &     9.90 \\
        \quad + BioBLP-M  & \checkmark &     8.72 &     6.22 &     9.42 &    13.10 \\
        \quad + BioBLP-D  &            &     7.42 &     4.91 &     7.39 &    11.62 \\
        \quad + BioBLP-D  & \checkmark &    15.74 &    11.73 &    16.44 &    23.04 \\
        \midrule
        ComplEx           &            &    42.75 &    31.55 &    48.09 &    65.50 \\
        \quad + BioBLP-P  &            &    16.53 &    11.13 &    17.17 &    26.92 \\
        \quad + BioBLP-P  & \checkmark &    34.88 &    25.11 &    39.52 &    54.83 \\
        \quad + BioBLP-M  &            &     1.68 &     1.24 &     1.86 &     2.49 \\
        \quad + BioBLP-M  & \checkmark &     1.78 &     1.37 &     1.93 &     2.54 \\
        \quad + BioBLP-D  &            &     0.01 &     0.00 &     0.00 &     0.02 \\
        \quad + BioBLP-D  & \checkmark &     0.45 &     0.36 &     0.43 &     0.55 \\
        \midrule
        RotatE            &            &    55.20 &    44.46 &    61.95 &    74.76 \\
        \quad + BioBLP-P  &            &    45.29 &    35.60 &    51.33 &    62.89 \\
        \quad + BioBLP-P  & \checkmark &    47.30 &    36.56 &    54.59 &    66.10 \\
        \quad + BioBLP-M  &            &    10.40 &     7.02 &    10.40 &    16.30 \\
        \quad + BioBLP-M  & \checkmark &    14.34 &    11.14 &    14.79 &    19.78 \\
        \quad + BioBLP-D  &            &    11.60 &     8.79 &    12.43 &    16.67 \\
        \quad + BioBLP-D  & \checkmark &    49.68 &    40.62 &    55.30 &    66.26 \\
        \bottomrule
    \end{tabular}
\end{table}

The results are shown in Table~\ref{tab:link-prediction-results}, where we compare BioBLP with our baselines.
We observe that in all cases, introducing different kinds of attribute encoders results in lower performance when compared to the baselines that rely on the structure of the graph alone to learn embeddings.
However, we note that when combined with the RotatE scoring function, BioBLP yields competitive performance when using protein and disease encoders.
This implies a trade-off to be considered when learning embeddings from attribute data: while lookup table embedding methods are effective at link prediction, they are limited to the set of entities observed during training.
BioBLP results in a drop in performance, but it provides a more general embedding method that can generate representations for entities outside the training set of entities.

As observed during the hyperparameter study, RotatE results in the best performance across all models.
Its performance also transfers to BioBLP models that use the RotatE scoring function, indicating that regardless of the mechanism used for embedding entities (be it lookup tables, or attribute encoders), the geometry of the space induced by RotatE and its scoring function are particularly effective for link prediction on BioKG.
When using BioBLP with other scoring functions, performance is significantly lower.
Notably, ComplEx+BioBLP-D fails at solving the link prediction task.

When focusing on the best-performing group of models that use RotatE, we can see that the lowest performance is obtained when using molecule encoders, suggesting that further improvements are required in how molecules are represented an encoded in BioBLP-M models.
When using protein and disease encoders, link prediction performance increases, with RotatE+BioBLP-D yielding the highest performance over all our models that use encoders.

\begin{figure}
    \centering
    \includegraphics[width=0.95\linewidth]{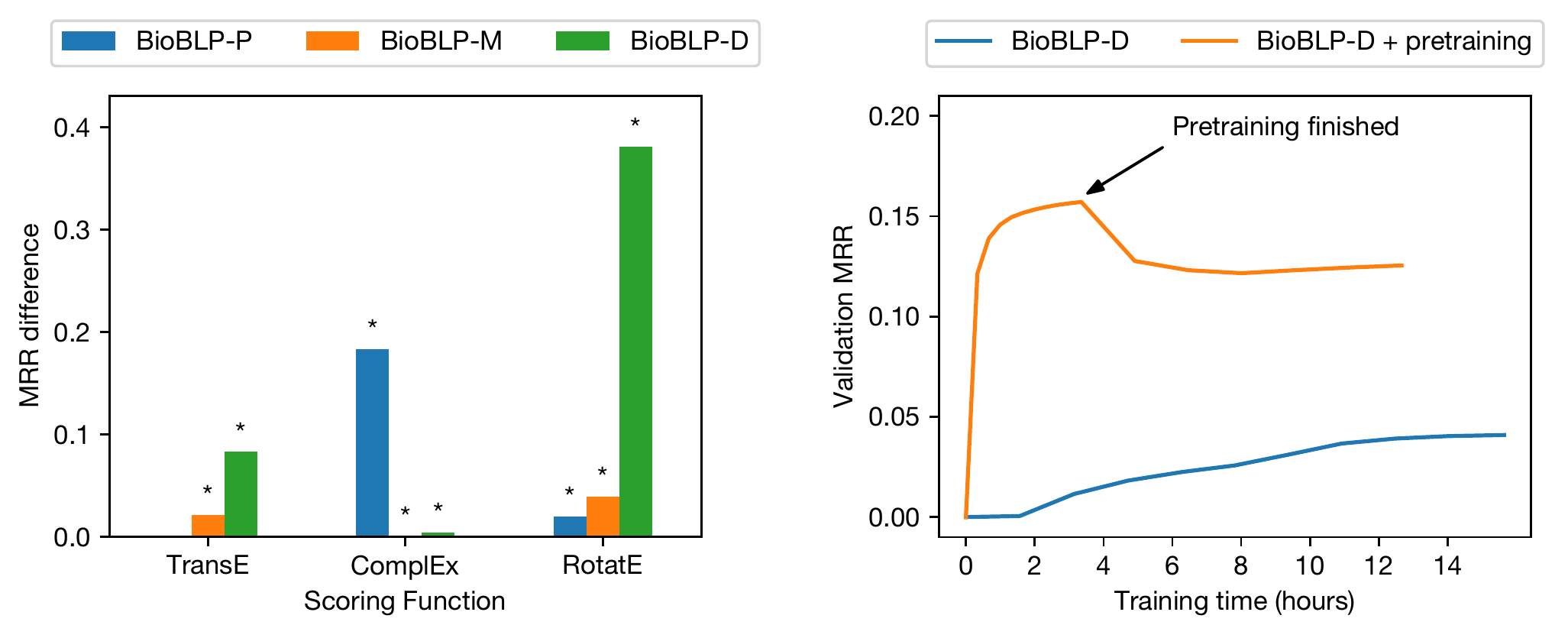}
    \caption{The impact of our proposed pretraining strategy. Left: increase in MRR due to the pretraining strategy over base models trained from scratch. Asterisks indicate significance at p-values less than 0.05. Right: A comparison of the runtime required until convergence on the validation set, when training BioBLP-D from scratch, and when including pretraining. During pretraining, attributes are ignored, and after pretraining we start optimizing the attribute encoders.}
    \label{fig:pretraining-significance}
\end{figure}

We note that our proposed strategy for pretraining always increases the performance over the corresponding models that have not been pretrained.
The magnitude of such improvements varies, with some resulting in slightly higher performance and others displaying pronounced differences.
The best variant that we obtained corresponds to a pretrained RotatE+BioBLP-D model, whose performance increases by approximately four times when compared to the model that is not pretrained.
We test for significance using a paired Welch's $t$-test by comparing the reciprocal rank per triple in the test set, between models trained from scratch and pretrained models.
We illustrate the impact of pretraining in Fig.~\ref{fig:pretraining-significance}, where we plot the increase in MRR, and the performance until convergence as a function of time.
With the exception of BioBLP-P with the TransE scoring function, we find that the increase in MRR provided by pretraining is statistically significant (p $<$ 0.05).
We also note that pretraining reduces the time required to converge and reach higher link prediction performance, compared to no pretraining.
Even when allocating more time resources to the setting without pretraining, the model reaches much lower values of MRR.

\subsection{Benchmarks}
\label{sec:results-benchmarks}

We present results on the DPI benchmarks using the BioBLP models that employ the RotatE scoring function, since these provided significantly higher performance in the link prediction task compared to variants using other scoring functions.

\begin{figure}[t]
    \centering
    \includegraphics[width=0.95\linewidth]{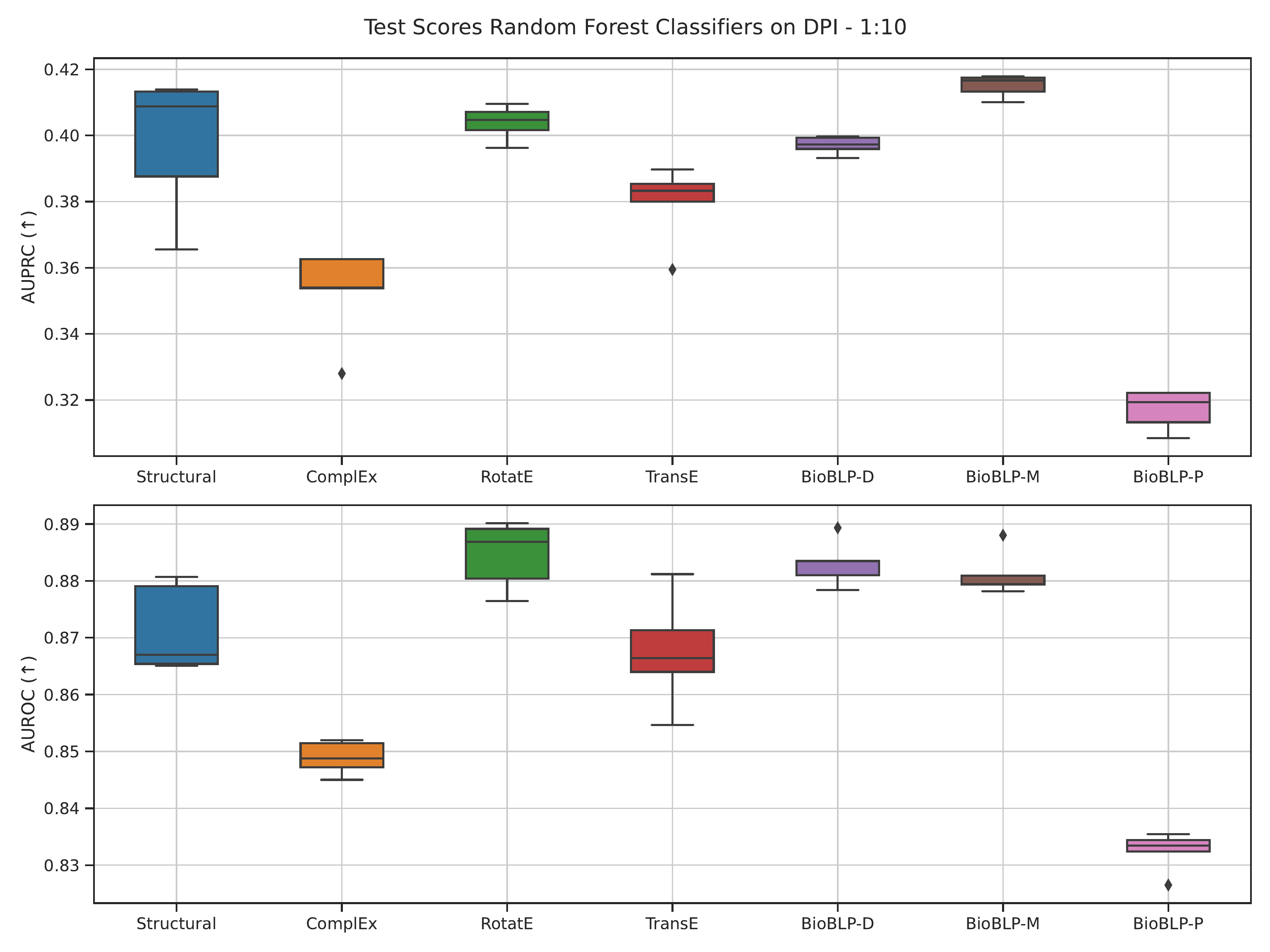}
    \caption{Classification performance for Random Forest Classifiers for DPI benchmark with positive-negative ratio of 1:10. The box for the baseline classifier using Random embeddings is omitted for clarity, reaching mean AUPRC of $0.192$ and mean AUROC of $0.720$. The median is indicated by the horizontal line, the quartiles indicated by the box, the min and max by the whiskers and individual points are outliers.}
    \label{fig:dpi-scores-r10}
\end{figure}

In Figure~\ref{fig:dpi-scores-r10} we report the AUROC and AUPRC metrics for the RF classifiers, which resulted on better performance compared to LR and MLP classifiers, on the five test folds for positive-negative ratio 1:10. 
We focus on these models to discuss the impact of the type of embedding for solving the DPI task. 
We provide extended results for all classifiers in Appendix~\ref{app:benchmarks}.

We observe that classifiers trained on BioBLP embeddings remain competitive with the baseline embeddings considered in this work. 
The Random baseline (not shown in  Fig.~\ref{fig:dpi-scores-r10}) reached mean AUROC of $0.720$ and mean AUPRC of $0.192$.
All models score above these values, confirming the classifiers learn the task better than chance when given the proposed embeddings as features.
Among the baselines, RotatE and Structural embeddings lead to highest mean scores in the AUROC and AUPRC metrics.
The model trained on BioBLP-M reaches highest mean AUPRC overall, but on the AUROC metric scores below BioBLP-D and RotatE embeddings.
Notably, the models trained on BioBLP-P embeddings score lowest in this setup, while this model was among the top scoring models in the link prediction setup.

These results show that embeddings from BioBLP-M and BioBLP-D, which encode molecular structures and disease descriptions, provide competitive performance when compared to lookup table embeddings.
In contrast, embeddings from BioBLP-P, which encodes protein sequences, perform worse among other types of embeddings.
In particular, BioBLP-P performs worse than the Structural baseline, whose embeddings are obtained from ProtTrans and MolTrans.
Since we also employ ProtTrans embeddings as inputs for BioBLP-P, this indicates that using BioBLP-P \emph{degrades} the quality of the ProtTrans embeddings, thus further work is required in properly incorporating protein sequences into encoders for BioBLP.
We do not observe such degradation in BioBLP-M, which uses MolTrans embeddings as inputs but outperforms the Structural baseline.

\subsection{Performance analysis}

Lookup table embeddings such as RotatE have been observed to suffer from node-degree bias, obtaining high link prediction scores by learning to predict links for a small fraction of entities that are responsible for a large number of triples in the graph \cite{rossi2020knowledge}.
These models show poor link prediction performance in sparser regions of the graph comprised by entities with a low degree of incoming or outgoing edges.

We investigate the performance of link prediction models with respect to the degree of the entities being predicted, focusing on a comparison between RotatE and BioBLP-D, which uses an encoder of descriptions of diseases.
We select all test triples involving diseases, and we compute the difference in MRR between BioBLP-D and RotatE when i) given a non-disease entity and a relation, \emph{diseases are predicted}; and ii) given a disease entity and a relation, \emph{non-disease entities are predicted}.
A positive difference indicates that RotatE results in higher performance.

\begin{figure}[t]
\begin{subfigure}{0.48\textwidth}
  \centering
  \includegraphics[width=\linewidth]{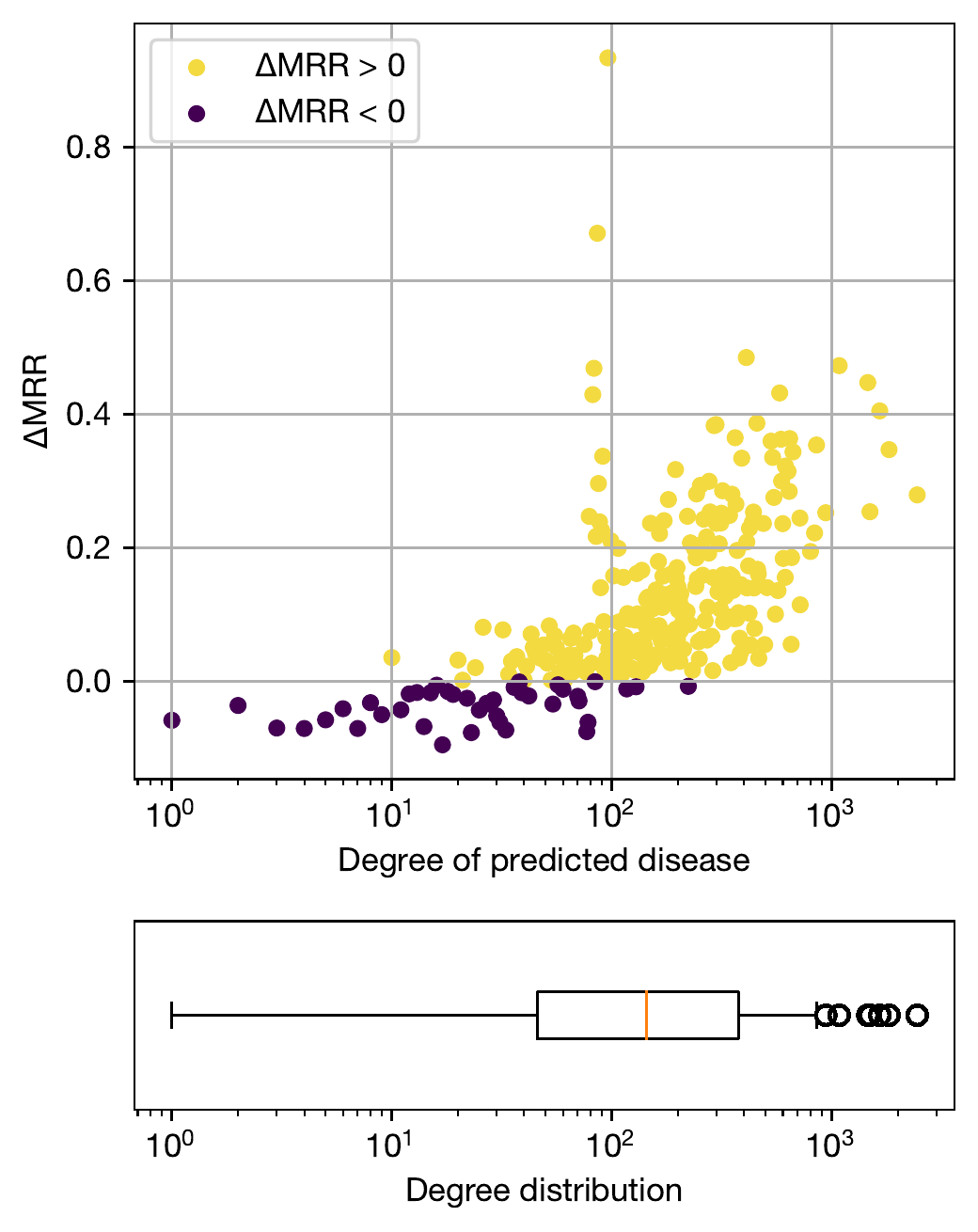}
   \caption{Predicting diseases}
  \label{fig:degree-analysis-disease}
\end{subfigure}%
\begin{subfigure}{0.49\textwidth}
  \centering
  \includegraphics[width=\linewidth]{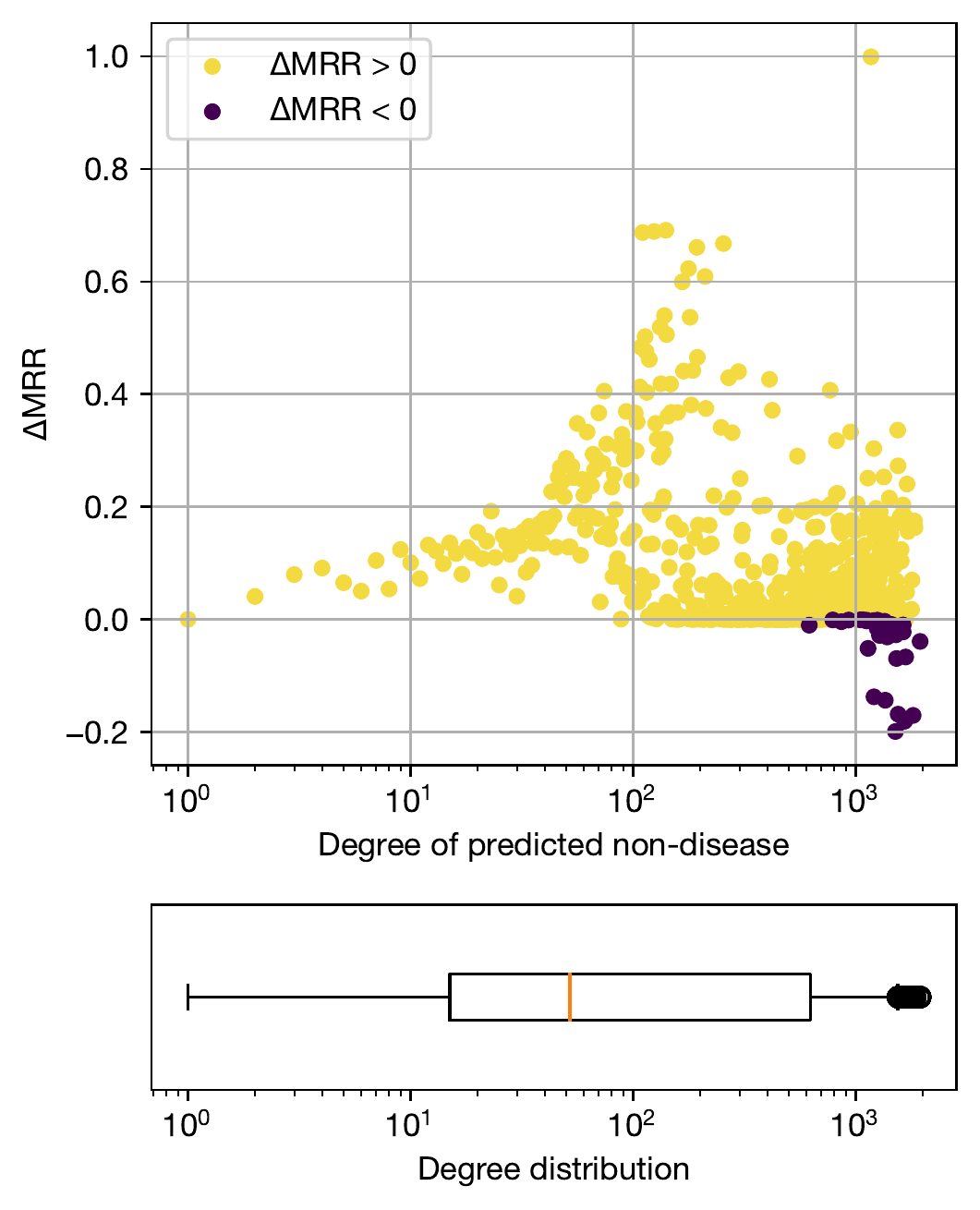}
  \caption{Predicting non-diseases}
  \label{fig:degree-analysis-nondisease}
\end{subfigure}
    \caption{Difference in Average MRR between RotatE and BioBLP-D predictions against node degree of \textit{disease} (Fig. (a)) or \textit{non-disease} (Fig. (b)) entity being predicted, together with the degree distribution of predicted entities. 
A negative $\Delta$MRR means that BioBLP-D outperforms RotatE, while a positive value means that RotatE has higher performance.}
    \label{fig:avg_mrr_vs_node_degree}
\end{figure}

The results are shown in Fig.~\ref{fig:avg_mrr_vs_node_degree}, where we present the difference in MRR, together with a visualization of the degree distribution for predicted entities.
From Fig.~\ref{fig:degree-analysis-disease}, we observe that BioBLP-D outperforms RotatE when the predicted entity is a disease, over a subset of low degree entities that range from a degree of 1 to approximately 20.
For entities with a degree between 20 and 100 we observed mixed results, with multiple cases in which BioBLP-D produces better results.
In the case of high degree entities, RotatE consistently performs better than BioBLP-D when predicting a disease.

When given a disease and predicting a non-disease, Fig.~\ref{fig:degree-analysis-nondisease} shows that RotatE works better across different degree values, with the exception of a small set of high-degree diseases where BioBLP-D results in higher MRR.

\begin{table}[t]
\caption{Quantiles for node degree per entity type in the BioKG training graph used in our work. 
The node degree of an entity is the sum of the incoming and outgoing edges from the entity.}
\label{tab:biokg-node-degree-stats}
\begin{tabular}{lcccccc}
\toprule
\textbf{Statistic} & \textbf{mean $\pm$ std} & \textbf{min} & \textbf{25\%} & \textbf{50\%} & \textbf{75\%} & \textbf{max}\\
\midrule
\quad Proteins     &       9.60 $\pm$  19.33 &            1 &             2 &             4 &             9 &  480        \\
\quad Diseases     &      25.89 $\pm$  85.37 &            1 &             3 &             5 &            15 & 2448        \\
\quad Drugs        &     218.85 $\pm$ 351.24 &            1 &             2 &             6 &           369 & 1942        \\ 
\quad All          &      27.95 $\pm$ 119.00 &            1 &             2 &             4 &            10 & 2448        \\ 
\bottomrule
\end{tabular}
\centering
\end{table}

The degree distributions visualized in Fig.~\ref{fig:avg_mrr_vs_node_degree} suggest that a substantial amount of entities in the graph have low to medium degrees, while entities of high degree are few.
This is expected in graphs where the distribution of the degree follows a power law, and it can occur in the biomedical domain where some entities are involved in multiple processes, while other entities have specific functions or are understudied~\cite{bonner2022topological}.
We confirm this by examining the quantiles of degree values for different entity types in Table~\ref{tab:biokg-node-degree-stats}.
We note that 50\% of the entities in the graph have degrees of up to 6, which is well below the degree values at which BioBLP-D outperforms RotatE.
Interestingly, this result is not apparent in the standard link prediction evaluation, where metrics are micro-averaged over the complete set of triples and are thus dominated by the performance over high-degree entities.

\section{Discussion}

With BioBLP, we address limitations of existing work on embedding biomedical KGs that assumes that all entities in the KG are associated with the same data modality or that such information is available for all entities in the graph.
We observe that when combined with the RotatE scoring function, BioBLP-D provided competitive, yet lower performance in the link prediction and DPI benchmark, in comparison with a RotatE baseline that uses lookup table embeddings and thus does not take attribute data into account.
This is in line with prior works that explore the use of textual descriptions to learn entity embeddings~\cite{daza2021inductive}.
Optimizing the molecule and protein encoders in BioBLP-M and BioBLP-P proved more challenging, resulting in embeddings with lower performance. 
Future work could attempt to find ways to create encoders for these modalities that are expressive while keeping training over large KGs scalable.

On one hand, these findings indicate that RotatE is an effective model for link prediction on BioKG, as well as a suitable scoring function when used in the BioBLP framework.
However, we note that the common evaluation protocol where link prediction metrics are micro-averaged over a test set of triples biases the results such that the average is dominated by high-degree entities in the graph.
This issue has been highlighted in prior work on evaluation protocols for link prediction methods~\cite{rossi2020knowledge}.
We argue that for tasks of scientific discovery, the performance over low degree entities is essential - if not more important - since these are understudied areas of the graph that would benefit more from predicting potential links that do not exist in the KG yet.
Our analysis shows that BioBLP-D achieves better performance than RotatE when predicting edges to low degree diseases, which comprise a substantial amount of the complete set of entities in the KG.
Based on this finding, we call for future scientific discovery work which develops alternative evaluation protocols for assessing link prediction methods in that context.

Further, we observed that training BioBLP results in increased computational costs, since the attribute encoders are deep neural networks with multiple layers and more parameters in comparison with the shallow, lookup table embeddings found in models like RotatE.
Our results show that pretraining is an effective strategy for lowering the computational cost of training BioBLP.
Pretraining is done using a model that is drastically cheaper to optimize, such as RotatE, so that once the relevant modules are transferred to BioBLP, less training steps are required and the resulting performance is higher compared to omitting pretraining.

A useful property of BioBLP is its ability to encode arbitrary entities, rather than entities predefined in the training set, as long as attribute data is available for them.
While our focus was comparing lookup table embeddings with BioBLP, future work could focus on the inductive setting where lookup table embeddings are not applicable.
Our framework could also be further extended by considering that a single entity might have not one but multiple attributes, such as SMILES representations \emph{and} molecular graphs.

The use of attribute encoders in BioBLP suggests a new interface for leveraging the information contained in a KG with attribute data.
One interesting application for our proposed approach is the development of a natural language interface with the KG for biomedical tasks.
Using the text encoder component, we have the ability to make predictions such as the effectiveness of a specific drug for a particular disease, even if the disease is not explicitly present in the KG.

Another use case that we would explore in future work is expanding the KG with new entities.
As KGs are inherently incomplete and require frequent updates, we could use BioBLP to compute embeddings for incoming entities, discover the closest entities by attribute similarity, and predict the most appropriate links with entities already in the KG.

\section{Conclusions}
We have presented BioBLP, a modular framework for learning embeddings of entities in a KG with attribute data, that also supports entities with missing attribute data.
We have additionally introduced a pretraining strategy for reducing training runtime.
Experiments on link prediction and domain-specific benchmarks show that BioBLP offers competitive performance, but further work is required to match the average performance of methods that do not use entity attributes.
In the specific case of low degree entities, that we argue are of special interest in the area of scientific discovery, we find settings where BioBLP is able to outperform such baselines.
In future work, we foresee the investigation of the tradeoff that arises when designing attribute encoders that are expressive yet scalable, as well as applications enabled by the use of multi-modal attribute encoders in biomedical knowledge graphs.

\backmatter

\section{Declarations}

\paragraph{Ethics approval and consent to participate}
Not applicable.

\paragraph{Consent for publication}
Not applicable.

\paragraph{Availability of data and materials}
The datasets generated and/or analysed during the current study are available in the Zenodo repository, available at \url{https://doi.org/10.5281/zenodo.8005711}. Our software implementation is open source and available at \url{https://github.com/elsevier-AI-Lab/BioBLP}.

\paragraph{Competing interests}
The authors declare that they have no competing interests.

\paragraph{Funding}
This project was funded by Elsevier’s Discovery Lab.

\paragraph{Authors' contributions}
D.D., D.A., P.M., and T.P. prepared and analyzed the data, designed models and experiments, and executed experiments. M.C. and P.G. supervised the project. All authors read and approved the manuscript.

\paragraph{Acknowledgements}
Not applicable.

\begin{appendices}

\section{Additional benchmark results}\label{app:benchmarks}

We present extended results on the DPI benchmark for all three classifiers (logistic regression, multi-layer perceptron, and random forest) on Table \ref{tab:results-benchmarks-all-r10}.
In most cases, we observe that the random forest outperforms the other classifiers on the DPI task when using KG embeddings as inputs.

\begin{table}[h]
\caption{Classification performance on DPI benchmark with ratio 1:10. Reported are the mean (std) over the 5 best models scored on the test folds. }
\label{tab:results-benchmarks-all-r10}
\centering
\begin{tabular}{llccc}
\toprule     
        &           &         \mc{3}{c}{DPI-FDA}                           \\ 
                        \cmidrule(lr){3-5}                                         
Feature    & Classifier &    AUPRC       &       AUROC    &     F1         \\
\midrule
Random     & LR         &     0.129 ± 0.003 &     0.597 ± 0.001 &     0.602 ± 0.010 \\
           & MLP        &     0.139 ± 0.006 &     0.635 ± 0.014 &     0.594 ± 0.004 \\
           & RF         & \bf 0.192 ± 0.004 & \bf 0.720 ± 0.005 & \bf 0.889 ± 0.001 \\
\midrule
Structural & LR         &     0.153 ± 0.003 &     0.690 ± 0.010 &     0.617 ± 0.010 \\
           & MLP        &     0.244 ± 0.006 &     0.771 ± 0.003 &     0.665 ± 0.010 \\
           & RF         & \bf 0.398 ± 0.021 & \bf 0.871 ± 0.008 & \bf 0.885 ± 0.016 \\
\midrule
ComplEx    & LR         &     0.170 ± 0.002 &     0.706 ± 0.004 &     0.639 ± 0.004 \\
           & MLP        &     0.339 ± 0.006 &     0.838 ± 0.004 &     0.755 ± 0.003 \\
           & RF         & \bf 0.352 ± 0.014 & \bf 0.849 ± 0.003 & \bf 0.892 ± 0.008 \\
\midrule
RotatE     & LR         &     0.180 ± 0.002 &     0.722 ± 0.005 &     0.643 ± 0.003 \\
           & MLP        & \bf 0.453 ± 0.008 & \bf 0.886 ± 0.002 &     0.794 ± 0.006 \\
           & RF         &     0.404 ± 0.005 &     0.885 ± 0.006 & \bf 0.905 ± 0.006 \\
\midrule
TransE     & LR         &     0.170 ± 0.003 &     0.691 ± 0.021 &     0.650 ± 0.017 \\
           & MLP        &     0.350 ± 0.009 &     0.846 ± 0.005 &     0.750 ± 0.003 \\
           & RF         & \bf 0.380 ± 0.012 & \bf 0.868 ± 0.010 & \bf 0.894 ± 0.015 \\
\midrule
BioBLP-D   & LR         &     0.177 ± 0.003 &     0.723 ± 0.003 &     0.641 ± 0.002 \\
           & MLP        & \bf 0.447 ± 0.008 & \bf 0.885 ± 0.003 &     0.792 ± 0.002 \\
           & RF         &     0.397 ± 0.003 &     0.883 ± 0.004 & \bf 0.907 ± 0.004 \\
\midrule
BioBLP-M   & LR         &     0.166 ± 0.002 &     0.713 ± 0.002 &     0.634 ± 0.003 \\
           & MLP        &     0.401 ± 0.007 &     0.865 ± 0.002 &     0.764 ± 0.002 \\
           & RF         & \bf 0.415 ± 0.003 & \bf 0.881 ± 0.004 & \bf 0.895 ± 0.006 \\
\midrule
BioBLP-P   & LR         &     0.170 ± 0.006 &     0.683 ± 0.005 &     0.636 ± 0.005 \\
           & MLP        & \bf 0.343 ± 0.008 & \bf 0.820 ± 0.005 &     0.744 ± 0.015 \\
           & RF         &     0.317 ± 0.006 &     0.832 ± 0.004 & \bf 0.899 ± 0.002 \\
\midrule
\bottomrule
\end{tabular}
\end{table}

\end{appendices}

\bibliography{references}

\begin{thebibliography}{10}
\providecommand{\doi}[1]{\url{https://doi.org/#1}}
\bibcommenthead

\bibitem[\protect\citeauthoryear{Ritchie}{2018}]{Ritchie2018}
Ritchie MD.
\newblock Large-Scale Analysis of Genetic and Clinical Patient Data.
\newblock Annual Review of Biomedical Data Science. 2018 Jul;1(1):263--274.

\bibitem[\protect\citeauthoryear{Stephens et~al.}{2015}]{stephens2015big}
Stephens ZD, et~al.
\newblock Big {Data}: astronomical or genomical?
\newblock PLoS biology. 2015;13(7):e1002195.

\bibitem[\protect\citeauthoryear{Zhu}{2020}]{Zhu2020}
Zhu H.
\newblock {Big Data} and Artificial Intelligence Modeling for Drug Discovery.
\newblock Annual Review of Pharmacology and Toxicology. 2020
  Jan;60(1):573--589.

\bibitem[\protect\citeauthoryear{Wilkinson et~al.}{2016}]{Wilkinson2016}
Wilkinson MD, Dumontier M, et~al.
\newblock {The {FAIR} Guiding Principles for scientific data management and
  stewardship}.
\newblock Scientific Data. 2016;3(1):160018.

\bibitem[\protect\citeauthoryear{Waagmeester et~al.}{2020}]{Waagmeester2020}
Waagmeester A, Stupp G, et~al.
\newblock Wikidata as a knowledge graph for the life sciences.
\newblock eLife. 2020 mar;9.

\bibitem[\protect\citeauthoryear{Belleau et~al.}{2008}]{belleau2008bio2rdf}
Belleau F, Nolin MA, Tourigny N, Rigault P, Morissette J.
\newblock {Bio2RDF}: towards a mashup to build bioinformatics knowledge
  systems.
\newblock Journal of biomedical informatics. 2008;41(5):706--716.

\bibitem[\protect\citeauthoryear{Williams et~al.}{2012}]{williams2012open}
Williams AJ, Harland L, Groth P, Pettifer S, Chichester C, Willighagen EL,
  et~al.
\newblock Open PHACTS: semantic interoperability for drug discovery.
\newblock Drug discovery today. 2012;17(21-22):1188--1198.

\bibitem[\protect\citeauthoryear{Domingo-Fernández
  et~al.}{2020}]{covid19kg2020}
Domingo-Fernández D, Baksi S, et~al.
\newblock {{COVID-19} Knowledge Graph: a computable, multi-modal,
  cause-and-effect knowledge model of {COVID-19} pathophysiology}.
\newblock Bioinformatics. 2020 09;.

\bibitem[\protect\citeauthoryear{Himmelstein et~al.}{2017}]{Himmelstein2017}
Himmelstein DS, Lizee A, et~al.
\newblock Systematic integration of biomedical knowledge prioritizes drugs for
  repurposing.
\newblock eLife. 2017 sep;6:e26726.

\bibitem[\protect\citeauthoryear{Hogan et~al.}{2021}]{kg-book}
Hogan A, Blomqvist E, Cochez M, d'Amato C, de~Melo G, Guti\'errez C, et~al.
\newblock {K}nowledge {G}raphs.
\newblock No.~22 in Synthesis Lectures on Data, Semantics, and Knowledge.
  Springer; 2021.
\newblock Available from: \url{https://kgbook.org/}.

\bibitem[\protect\citeauthoryear{Chichester et~al.}{2015}]{CHICHESTER2015399}
Chichester C, Digles D, et~al.
\newblock Drug discovery {FAQs}: workflows for answering multidomain drug
  discovery questions.
\newblock Drug Discovery Today. 2015;20(4):399--405.

\bibitem[\protect\citeauthoryear{Knox et~al.}{2010}]{knox2010drugbank}
Knox C, Law V, Jewison T, Liu P, et~al.
\newblock {DrugBank} 3.0: a Comprehensive Resource for `omics' Research on
  Drugs.
\newblock Nucleic acids research. 2010;39(suppl\_1):D1035--D1041.

\bibitem[\protect\citeauthoryear{Bateman et~al.}{2015}]{uniprot2015uniprot}
Bateman A, Martin MJ, O'Donovan C, Magrane M, Apweiler R, Alpi E, et~al.
\newblock {{U}ni{P}rot: a hub for protein information}.
\newblock Nucleic Acids Res. 2015 Jan;43(Database issue):D204--212.

\bibitem[\protect\citeauthoryear{Lipscomb}{2000}]{lipscomb2000medical}
Lipscomb CE.
\newblock Medical Subject Headings ({MeSH}).
\newblock Bull Med Libr Assoc. 2000;88(3): 265–266.

\bibitem[\protect\citeauthoryear{Morselli~Gysi
  et~al.}{2021}]{morselli2021network}
Morselli~Gysi D, Do~Valle {\'I}, Zitnik M, Ameli A, Gan X, Varol O, et~al.
\newblock Network medicine framework for identifying drug-repurposing
  opportunities for COVID-19.
\newblock Proceedings of the National Academy of Sciences.
  2021;118(19):e2025581118.

\bibitem[\protect\citeauthoryear{Nickel et~al.}{2016}]{nickel2016review}
Nickel M, Murphy K, Tresp V, Gabrilovich E.
\newblock A Review of Relational Machine Learning for Knowledge Graphs.
\newblock Proc {IEEE}. 2016;104(1):11--33.
\newblock \doi{10.1109/JPROC.2015.2483592}.

\bibitem[\protect\citeauthoryear{Wang et~al.}{2017}]{wang2017kgsurvey}
Wang Q, Mao Z, Wang B, Guo L.
\newblock Knowledge Graph Embedding: {A} Survey of Approaches and Applications.
\newblock {IEEE} Trans Knowl Data Eng. 2017;29(12):2724--2743.
\newblock \doi{10.1109/TKDE.2017.2754499}.

\bibitem[\protect\citeauthoryear{Bordes et~al.}{2013}]{bordes2013transe}
Bordes A, Usunier N, Garc{\'{\i}}a{-}Dur{\'{a}}n A, Weston J, Yakhnenko O.
\newblock Translating Embeddings for Modeling Multi-relational Data.
\newblock In: Burges CJC, Bottou L, Ghahramani Z, Weinberger KQ, editors.
  Advances in Neural Information Processing Systems 26: 27th Annual Conference
  on Neural Information Processing Systems 2013. Proceedings of a meeting held
  December 5-8, 2013, Lake Tahoe, Nevada, United States; 2013. p. 2787--2795.
\newblock Available from:
  \url{https://proceedings.neurips.cc/paper/2013/hash/1cecc7a77928ca8133fa24680a88d2f9-Abstract.html}.

\bibitem[\protect\citeauthoryear{Trouillon et~al.}{2016}]{trouillon2016complex}
Trouillon T, Welbl J, Riedel S, Gaussier {\'{E}}, Bouchard G.
\newblock Complex Embeddings for Simple Link Prediction.
\newblock In: Balcan M, Weinberger KQ, editors. Proceedings of the 33nd
  International Conference on Machine Learning, {ICML} 2016, New York City, NY,
  USA, June 19-24, 2016. vol.~48 of {JMLR} Workshop and Conference Proceedings.
  JMLR.org; 2016. p. 2071--2080.
\newblock Available from:
  \url{http://proceedings.mlr.press/v48/trouillon16.html}.

\bibitem[\protect\citeauthoryear{Sun et~al.}{2019}]{sun2019rotate}
Sun Z, Deng Z, Nie J, Tang J.
\newblock RotatE: Knowledge Graph Embedding by Relational Rotation in Complex
  Space.
\newblock In: 7th International Conference on Learning Representations, {ICLR}
  2019, New Orleans, LA, USA, May 6-9, 2019. OpenReview.net; 2019. Available
  from: \url{https://openreview.net/forum?id=HkgEQnRqYQ}.

\bibitem[\protect\citeauthoryear{Xie et~al.}{2016}]{xie2016dkrl}
Xie R, Liu Z, Jia J, Luan H, Sun M.
\newblock Representation Learning of Knowledge Graphs with Entity Descriptions.
\newblock In: Schuurmans D, Wellman MP, editors. Proceedings of the Thirtieth
  {AAAI} Conference on Artificial Intelligence, February 12-17, 2016, Phoenix,
  Arizona, {USA}. {AAAI} Press; 2016. p. 2659--2665.
\newblock Available from:
  \url{http://www.aaai.org/ocs/index.php/AAAI/AAAI16/paper/view/12216}.

\bibitem[\protect\citeauthoryear{Teru et~al.}{2020}]{teru2020inductive}
Teru KK, Denis EG, Hamilton WL.
\newblock Inductive Relation Prediction by Subgraph Reasoning.
\newblock In: Proceedings of the 37th International Conference on Machine
  Learning, {ICML} 2020, 13-18 July 2020, Virtual Event. vol. 119 of
  Proceedings of Machine Learning Research. {PMLR}; 2020. p. 9448--9457.
\newblock Available from: \url{http://proceedings.mlr.press/v119/teru20a.html}.

\bibitem[\protect\citeauthoryear{Daza et~al.}{2021}]{daza2021inductive}
Daza D, Cochez M, Groth P.
\newblock Inductive Entity Representations from Text via Link Prediction.
\newblock In: Leskovec J, Grobelnik M, Najork M, Tang J, Zia L, editors. {WWW}
  '21: The Web Conference 2021, Virtual Event / Ljubljana, Slovenia, April
  19-23, 2021. {ACM} / {IW3C2}; 2021. p. 798--808.
\newblock Available from: \url{https://doi.org/10.1145/3442381.3450141}.

\bibitem[\protect\citeauthoryear{Galkin et~al.}{2022}]{galkin2022nodepiece}
Galkin M, Denis EG, Wu J, Hamilton WL.
\newblock NodePiece: Compositional and Parameter-Efficient Representations of
  Large Knowledge Graphs.
\newblock In: The Tenth International Conference on Learning Representations,
  {ICLR} 2022, Virtual Event, April 25-29, 2022. OpenReview.net; 2022. p.
  1--14.
\newblock Available from: \url{https://openreview.net/forum?id=xMJWUKJnFSw}.

\bibitem[\protect\citeauthoryear{Xie et~al.}{2017}]{xie2017image}
Xie R, Liu Z, Luan H, Sun M.
\newblock Image-embodied Knowledge Representation Learning.
\newblock In: Sierra C, editor. Proceedings of the Twenty-Sixth International
  Joint Conference on Artificial Intelligence, {IJCAI} 2017, Melbourne,
  Australia, August 19-25, 2017. ijcai.org; 2017. p. 3140--3146.
\newblock Available from: \url{https://doi.org/10.24963/ijcai.2017/438}.

\bibitem[\protect\citeauthoryear{Tay et~al.}{2017}]{tay2017multitask}
Tay Y, Tuan LA, Phan MC, Hui SC.
\newblock Multi-Task Neural Network for Non-discrete Attribute Prediction in
  Knowledge Graphs.
\newblock In: Lim E, Winslett M, Sanderson M, Fu AW, Sun J, Culpepper JS,
  et~al., editors. Proceedings of the 2017 {ACM} on Conference on Information
  and Knowledge Management, {CIKM} 2017, Singapore, November 06 - 10, 2017.
  {ACM}; 2017. p. 1029--1038.
\newblock Available from: \url{https://doi.org/10.1145/3132847.3132937}.

\bibitem[\protect\citeauthoryear{Wu and Wang}{2018}]{wang2018numericembedding}
Wu Y, Wang Z.
\newblock Knowledge Graph Embedding with Numeric Attributes of Entities.
\newblock In: Proceedings of the Third Workshop on Representation Learning for
  {NLP}. Melbourne, Australia: Association for Computational Linguistics; 2018.
  p. 132--136.
\newblock Available from: \url{https://aclanthology.org/W18-3017}.

\bibitem[\protect\citeauthoryear{Pezeshkpour et~al.}{2018}]{pouya2018embedding}
Pezeshkpour P, Chen L, Singh S.
\newblock Embedding Multimodal Relational Data for Knowledge Base Completion.
\newblock In: Proceedings of the 2018 Conference on Empirical Methods in
  Natural Language Processing. Brussels, Belgium: Association for Computational
  Linguistics; 2018. p. 3208--3218.
\newblock Available from: \url{https://aclanthology.org/D18-1359}.

\bibitem[\protect\citeauthoryear{Kristiadi
  et~al.}{2019}]{kristiadi2019literale}
Kristiadi A, Khan MA, Lukovnikov D, Lehmann J, Fischer A.
\newblock Incorporating Literals into Knowledge Graph Embeddings.
\newblock In: Ghidini C, Hartig O, Maleshkova M, Sv{\'{a}}tek V, Cruz IF, Hogan
  A, et~al., editors. The Semantic Web - {ISWC} 2019 - 18th International
  Semantic Web Conference, Auckland, New Zealand, October 26-30, 2019,
  Proceedings, Part {I}. vol. 11778 of Lecture Notes in Computer Science.
  Springer; 2019. p. 347--363.
\newblock Available from: \url{https://doi.org/10.1007/978-3-030-30793-6\_20}.

\bibitem[\protect\citeauthoryear{Wang et~al.}{2021}]{wang2021kepler}
Wang X, Gao T, Zhu Z, Zhang Z, Liu Z, Li J, et~al.
\newblock {KEPLER:} {A} Unified Model for Knowledge Embedding and Pre-trained
  Language Representation.
\newblock Trans Assoc Comput Linguistics. 2021;9:176--194.
\newblock \doi{10.1162/tacl\_a\_00360}.

\bibitem[\protect\citeauthoryear{Ektefaie
  et~al.}{2023}]{ektefaie2023multimodal}
Ektefaie Y, Dasoulas G, Noori A, Farhat M, Zitnik M.
\newblock Multimodal learning with graphs.
\newblock Nature Machine Intelligence. 2023
  Apr;\doi{10.1038/s42256-023-00624-6}.

\bibitem[\protect\citeauthoryear{Wang et~al.}{2022}]{wang2022simkgc}
Wang L, Zhao W, Wei Z, Liu J.
\newblock SimKGC: Simple Contrastive Knowledge Graph Completion with
  Pre-trained Language Models.
\newblock In: Muresan S, Nakov P, Villavicencio A, editors. Proceedings of the
  60th Annual Meeting of the Association for Computational Linguistics (Volume
  1: Long Papers), {ACL} 2022, Dublin, Ireland, May 22-27, 2022. Association
  for Computational Linguistics; 2022. p. 4281--4294.
\newblock Available from: \url{https://doi.org/10.18653/v1/2022.acl-long.295}.

\bibitem[\protect\citeauthoryear{Markowitz et~al.}{2022}]{markowitz2022statik}
Markowitz E, Balasubramanian K, Mirtaheri M, Annavaram M, Galstyan A, Steeg GV.
\newblock StATIK: Structure and Text for Inductive Knowledge Graph Completion.
\newblock In: Carpuat M, de~Marneffe M, Ru{\'{\i}}z IVM, editors. Findings of
  the Association for Computational Linguistics: {NAACL} 2022, Seattle, WA,
  United States, July 10-15, 2022. Association for Computational Linguistics;
  2022. p. 604--615.
\newblock Available from:
  \url{https://doi.org/10.18653/v1/2022.findings-naacl.46}.

\bibitem[\protect\citeauthoryear{Safavi et~al.}{2022}]{safavi2022cascader}
Safavi T, Downey D, Hope T.
\newblock CascadER: Cross-Modal Cascading for Knowledge Graph Link Prediction.
\newblock CoRR. 2022;abs/2205.08012.
\newblock \doi{10.48550/arXiv.2205.08012}.
\newblock {\href{https://arxiv.org/abs/2205.08012}{{2205.08012}}}.

\bibitem[\protect\citeauthoryear{Devlin et~al.}{2019}]{devlin2019bert}
Devlin J, Chang M, Lee K, Toutanova K.
\newblock {BERT:} Pre-training of Deep Bidirectional Transformers for Language
  Understanding.
\newblock In: Burstein J, Doran C, Solorio T, editors. Proceedings of the 2019
  Conference of the North American Chapter of the Association for Computational
  Linguistics: Human Language Technologies, {NAACL-HLT} 2019, Minneapolis, MN,
  USA, June 2-7, 2019, Volume 1 (Long and Short Papers). Association for
  Computational Linguistics; 2019. p. 4171--4186.
\newblock Available from: \url{https://doi.org/10.18653/v1/n19-1423}.

\bibitem[\protect\citeauthoryear{Ali et~al.}{2019}]{ali2019biokeen}
Ali M, Hoyt CT, ndez D, Lehmann J, Jabeen H.
\newblock {{B}io{K}{E}{E}{N}: a library for learning and evaluating biological
  knowledge graph embeddings}.
\newblock Bioinformatics. 2019 Sep;35(18):3538--3540.

\bibitem[\protect\citeauthoryear{Nelson et~al.}{2019}]{nelson2019embed}
Nelson W, Zitnik M, Wang B, Leskovec J, Goldenberg A, Sharan R.
\newblock To embed or not: network embedding as a paradigm in computational
  biology.
\newblock Frontiers in genetics. 2019;10:381.

\bibitem[\protect\citeauthoryear{Walsh et~al.}{2020}]{walsh2020biokg}
Walsh B, Mohamed SK, Nov{\'a}{\v{c}}ek V.
\newblock Bio{KG}: A knowledge graph for relational learning on biological
  data.
\newblock In: Proceedings of the 29th ACM International Conference on
  Information \& Knowledge Management; 2020. p. 3173--3180.

\bibitem[\protect\citeauthoryear{Mohamed et~al.}{2020}]{mohamed2020trimodel}
Mohamed SK, ek V, Nounu A.
\newblock {{D}iscovering protein drug targets using knowledge graph
  embeddings}.
\newblock Bioinformatics. 2020 Jan;36(2):603--610.

\bibitem[\protect\citeauthoryear{Alshahrani
  et~al.}{2021}]{alshahrani2021application}
Alshahrani M, Thafar MA, Essack M.
\newblock {{A}pplication and evaluation of knowledge graph embeddings in
  biomedical data}.
\newblock PeerJ Comput Sci. 2021;7:e341.

\bibitem[\protect\citeauthoryear{Ye et~al.}{2022}]{ye2022tensorfac}
Ye C, Swiers R, Bonner S, Barrett I.
\newblock {{A} {K}nowledge {G}raph-{E}nhanced {T}ensor {F}actorisation {M}odel
  for {D}iscovering {D}rug {T}argets}.
\newblock IEEE/ACM Trans Comput Biol Bioinform. 2022 Aug;PP.

\bibitem[\protect\citeauthoryear{Gema et~al.}{2022}]{gema2023knowledge}
Gema AP, Grabarczyk D, De~Wulf W, Borole P, Alfaro JA, Minervini P, et~al.
\newblock Knowledge Graph Embeddings in the Biomedical Domain: Are They Useful?
  A Look at Link Prediction, Rule Learning, and Downstream Polypharmacy Tasks.
\newblock CoRR. 2022;abs/2305.19979.
\newblock \doi{10.48550/arXiv.2305.19979}.
\newblock {\href{https://arxiv.org/abs/2305.19979}{{2305.19979}}}.

\bibitem[\protect\citeauthoryear{Karim et~al.}{2019}]{karim2019drug}
Karim MR, Cochez M, Jares JB, Uddin M, Beyan OD, Decker S.
\newblock Drug-Drug Interaction Prediction Based on Knowledge Graph Embeddings
  and Convolutional-LSTM Network.
\newblock In: Shi XM, Buck M, Ma J, Veltri P, editors. Proceedings of the 10th
  {ACM} International Conference on Bioinformatics, Computational Biology and
  Health Informatics, {BCB} 2019, Niagara Falls, NY, USA, September 7-10, 2019.
  {ACM}; 2019. p. 113--123.
\newblock Available from: \url{https://doi.org/10.1145/3307339.3342161}.

\bibitem[\protect\citeauthoryear{Rossi and Matinata}{2020}]{rossi2020knowledge}
Rossi A, Matinata A.
\newblock Knowledge graph embeddings: Are relation-learning models learning
  relations?
\newblock In: EDBT/ICDT Workshops; 2020. .

\bibitem[\protect\citeauthoryear{Choi and Lee}{2021}]{choi2021identifying}
Choi W, Lee H.
\newblock Identifying disease-gene associations using a convolutional neural
  network-based model by embedding a biological knowledge graph with entity
  descriptions.
\newblock Plos one. 2021;16(10):e0258626.

\bibitem[\protect\citeauthoryear{Alshahrani
  et~al.}{2022}]{alshahrani2022combining}
Alshahrani M, Almansour A, Alkhaldi A, Thafar MA, Uludag M, Essack M, et~al.
\newblock {{C}ombining biomedical knowledge graphs and text to improve
  predictions for drug-target interactions and drug-indications}.
\newblock PeerJ. 2022;10:e13061.

\bibitem[\protect\citeauthoryear{Ren et~al.}{2022}]{ren2022biomedical}
Ren Z, You Z, Yu C, Li L, Guan Y, Guo L, et~al.
\newblock A biomedical knowledge graph-based method for drug-drug interactions
  prediction through combining local and global features with deep neural
  networks.
\newblock Briefings Bioinform. 2022;23(5).
\newblock \doi{10.1093/bib/bbac363}.

\bibitem[\protect\citeauthoryear{Su et~al.}{2022}]{su2022attention}
Su X, Hu L, You Z, Hu P, Zhao B.
\newblock Attention-based knowledge graph representation learning for
  predicting drug-drug interactions.
\newblock Briefings in bioinformatics. 2022;23(3):bbac140.

\bibitem[\protect\citeauthoryear{Zhang et~al.}{2022}]{zhang2022mkge}
Zhang Y, Li Z, Duan B, Qin L, Peng J.
\newblock {MKGE:} Knowledge graph embedding with molecular structure
  information.
\newblock Comput Biol Chem. 2022;100:107730.
\newblock \doi{10.1016/j.compbiolchem.2022.107730}.

\bibitem[\protect\citeauthoryear{Zhu et~al.}{2022}]{zhu2022multimodal}
Zhu C, Yang Z, Xia X, Li N, Zhong F, Liu L.
\newblock Multimodal reasoning based on knowledge graph embedding for specific
  diseases.
\newblock Bioinform. 2022;38(8):2235--2245.
\newblock \doi{10.1093/bioinformatics/btac085}.

\bibitem[\protect\citeauthoryear{Ruffinelli et~al.}{2020}]{ruffinelli2020dog}
Ruffinelli D, Broscheit S, Gemulla R.
\newblock You {CAN} Teach an Old Dog New Tricks! On Training Knowledge Graph
  Embeddings.
\newblock In: 8th International Conference on Learning Representations, {ICLR}
  2020, Addis Ababa, Ethiopia, April 26-30, 2020. OpenReview.net; 2020.
  Available from: \url{https://openreview.net/forum?id=BkxSmlBFvr}.

\bibitem[\protect\citeauthoryear{Ali et~al.}{2022}]{ali2022bringing}
Ali M, Berrendorf M, Hoyt CT, Vermue L, Galkin M, Sharifzadeh S, et~al.
\newblock Bringing Light Into the Dark: {A} Large-Scale Evaluation of Knowledge
  Graph Embedding Models Under a Unified Framework.
\newblock {IEEE} Trans Pattern Anal Mach Intell. 2022;44(12):8825--8845.
\newblock \doi{10.1109/TPAMI.2021.3124805}.

\bibitem[\protect\citeauthoryear{Bonner et~al.}{2022}]{bonner2022understanding}
Bonner S, Barrett IP, Ye C, Swiers R, Engkvist O, Hoyt CT, et~al.
\newblock Understanding the performance of knowledge graph embeddings in drug
  discovery.
\newblock Artificial Intelligence in the Life Sciences. 2022;2:100036.
\newblock \doi{https://doi.org/10.1016/j.ailsci.2022.100036}.

\bibitem[\protect\citeauthoryear{Wishart et~al.}{2008}]{wishart2008drugbank}
Wishart DS, Knox C, Guo AC, Cheng D, Shrivastava S, Tzur D, et~al.
\newblock {{D}rug{B}ank: a knowledgebase for drugs, drug actions and drug
  targets}.
\newblock Nucleic Acids Res. 2008 Jan;36(Database issue):D901--906.

\bibitem[\protect\citeauthoryear{Yamanishi
  et~al.}{2008}]{yamanishi2008prediction}
Yamanishi Y, Araki M, Gutteridge A, Honda W, Kanehisa M.
\newblock {{P}rediction of drug-target interaction networks from the
  integration of chemical and genomic spaces}.
\newblock Bioinformatics. 2008 Jul;24(13):i232--240.

\bibitem[\protect\citeauthoryear{Elnaggar et~al.}{2021}]{elnaggar2021prottrans}
Elnaggar A, Heinzinger M, Dallago C, Rehawi G, Wang Y, Jones L, et~al.
\newblock Prottrans: Toward understanding the language of life through
  self-supervised learning.
\newblock IEEE transactions on pattern analysis and machine intelligence.
  2021;44(10):7112--7127.

\bibitem[\protect\citeauthoryear{Morris et~al.}{2020}]{morris2020transformer}
Morris P, St~Clair R, Hahn WE, Barenholtz E.
\newblock Predicting Binding from Screening Assays with Transformer Network
  Embeddings.
\newblock Journal of Chemical Information and Modeling. 2020
  Jun;\doi{10.1021/acs.jcim.9b01212}.

\bibitem[\protect\citeauthoryear{Lee et~al.}{2019}]{jinhyuk2019biobert}
Lee J, Yoon W, Kim S, Kim D, Kim S, So CH, et~al.
\newblock {BioBERT: a pre-trained biomedical language representation model for
  biomedical text mining}.
\newblock Bioinformatics. 2019 09;36(4):1234--1240.
\newblock \doi{10.1093/bioinformatics/btz682}.
\newblock
  {\href{https://arxiv.org/abs/https://academic.oup.com/bioinformatics/article-pdf/36/4/1234/48983216/bioinformatics\_36\_4\_1234.pdf}{{https://academic.oup.com/bioinformatics/article-pdf/36/4/1234/48983216/bioinformatics\_36\_4\_1234.pdf}}}.

\bibitem[\protect\citeauthoryear{Vaswani et~al.}{2017}]{vaswani2017attention}
Vaswani A, Shazeer N, Parmar N, Uszkoreit J, Jones L, Gomez AN, et~al.
\newblock Attention is All you Need.
\newblock In: Guyon I, von Luxburg U, Bengio S, Wallach HM, Fergus R,
  Vishwanathan SVN, et~al., editors. Advances in Neural Information Processing
  Systems 30: Annual Conference on Neural Information Processing Systems 2017,
  December 4-9, 2017, Long Beach, CA, {USA}; 2017. p. 5998--6008.
\newblock Available from:
  \url{https://proceedings.neurips.cc/paper/2017/hash/3f5ee243547dee91fbd053c1c4a845aa-Abstract.html}.

\bibitem[\protect\citeauthoryear{Ali et~al.}{2021}]{ali2021pykeen}
Ali M, Berrendorf M, Hoyt CT, Vermue L, Sharifzadeh S, Tresp V, et~al.
\newblock PyKEEN 1.0: {A} Python Library for Training and Evaluating Knowledge
  Graph Embeddings.
\newblock J Mach Learn Res. 2021;22:82:1--82:6.

\bibitem[\protect\citeauthoryear{Nascimento
  et~al.}{2016}]{nascimento2016kronrls}
Nascimento AC, Prud{\^e}ncio RB, Costa IG.
\newblock A multiple kernel learning algorithm for drug-target interaction
  prediction.
\newblock BMC bioinformatics. 2016;17:1--16.

\bibitem[\protect\citeauthoryear{Hao et~al.}{2017}]{hao2017blmnii}
Hao M, Bryant SH, Wang Y.
\newblock Predicting drug-target interactions by dual-network integrated
  logistic matrix factorization.
\newblock Scientific reports. 2017;7(1):1--11.

\bibitem[\protect\citeauthoryear{Olayan et~al.}{2018}]{olayan2018ddr}
Olayan RS, Ashoor H, Bajic VB.
\newblock DDR: efficient computational method to predict drug--target
  interactions using graph mining and machine learning approaches.
\newblock Bioinformatics. 2018;34(7):1164--1173.

\bibitem[\protect\citeauthoryear{Takaya and Rehmsmeier}{2015}]{takaya2015}
Takaya MS, Rehmsmeier.
\newblock The Precision-Recall Plot Is More Informative than the ROC Plot When
  Evaluating Binary Classifiers on Imbalanced Datasets.
\newblock PLOS ONE. 2015 3;10:1--21.
\newblock \doi{10.1371/journal.pone.0118432}.

\bibitem[\protect\citeauthoryear{Bonner et~al.}{2022}]{bonner2022topological}
Bonner S, Kirik U, Engkvist O, Tang J, Barrett IP.
\newblock {Implications of topological imbalance for representation learning on
  biomedical knowledge graphs}.
\newblock Briefings in Bioinformatics. 2022 07;23(5).
\newblock Bbac279. \doi{10.1093/bib/bbac279}.
\newblock
  {\href{https://arxiv.org/abs/https://academic.oup.com/bib/article-pdf/23/5/bbac279/45937607/sup\_main\_bbac279.pdf}{{https://academic.oup.com/bib/article-pdf/23/5/bbac279/45937607/sup\_main\_bbac279.pdf}}}.

\end{thebibliography}

\end{document}